\documentclass[runningheads]{llncs}

\usepackage[table,dvipsnames]{xcolor}

\usepackage{eccv}
\usepackage{eccvabbrv}

\usepackage{tabularx}
\usepackage{makecell}
\usepackage{longtable}
\usepackage{microtype}
\usepackage{threeparttable}
\usepackage{threeparttablex}
\usepackage{graphicx}
\usepackage{booktabs}
\usepackage{marvosym}
\usepackage[accsupp]{axessibility} 
\usepackage{orcidlink}

\definecolor{dgray}{gray}{.5}

\usepackage{hyperref}

\begin{document}

% ---------------------------------------------------------------
% TODO REVIEW: Replace with your title
\title{Driver-WM: A Driver-Centric Traffic-Conditioned Latent World Model for In-Cabin Dynamics Rollout}

% TODO REVIEW: If the paper title is too long for the running head, you can set
% an abbreviated paper title here. If not, comment out.
% \titlerunning{Abbreviated paper title}
\titlerunning{Driver-WM}

% TODO FINAL: Replace with your author list. 
% Include the authors' OCRID for the camera-ready version, if at all possible.
% 放arxiv的时候解封，投稿ECCV得先不挂名
\author{Haozhuang Chi\inst{1}\orcidlink{0009-0000-7951-1046} \and
Daosheng Qiu\inst{2}\orcidlink{0009-0002-2855-5254} \and
Hao Su\inst{3}\orcidlink{0009-0004-7679-7842} \and
Haochen Liu\inst{1}\orcidlink{0000-0002-3628-8777} \and
Zirui Li\inst{1}\orcidlink{0000-0001-7056-4264} \and
Haoruo Zhang\inst{1}\orcidlink{0009-0007-6098-0943} \and
Chen Lv\inst{1}\textsuperscript{\Letter}\orcidlink{0000-0001-6897-4512}
}

% TODO FINAL: Replace with an abbreviated list of authors.
\authorrunning{H. Chi et al.} % 放arxiv的时候解封，投稿ECCV得先不挂名
% First names are abbreviated in the running head.
% If there are more than two authors, 'et al.' is used.

% ==========================================================
% TODO FINAL: 提交给 AUMOVIO 审核版本 / 最终 Camera-ready 录用版本 时使用
% ⚠️ 警告：3月5日提交给 ECCV 系统的盲审版本中，下边这段必须全部注释掉！保持目前的完全匿名状态！
% 应该写MAE还是NTU?我看最近几个arxiv是NTU
% \institute{Nanyang Technological University, Singapore\and
% Hubei University, Wuhan, China \and
% Osaka University, Osaka, Japan\\
% \email{Corresponding email: lyuchen@ntu.edu.sg}}

\institute{Nanyang Technological University, Singapore\and
Hubei University, Wuhan, China \and
Osaka University, Osaka, Japan\\
\email{Corresponding email: lyuchen@ntu.edu.sg}}
% ==========================================================

% 例子：
% TODO FINAL: Replace with your institution list.
% \institute{Princeton University, Princeton NJ 08544, USA \and
% Springer Heidelberg, Tiergartenstr.~17, 69121 Heidelberg, Germany
% \email{lncs@springer.com}\\
% \url{http://www.springer.com/gp/computer-science/lncs} \and
% ABC Institute, Rupert-Karls-University Heidelberg, Heidelberg, Germany\\
% \email{\{abc,lncs\}@uni-heidelberg.de}}

\maketitle

\begin{abstract}
Safe L2/L3 driving automation requires anticipating human-in-the-loop reactions during shared-control transitions. While most driving world models forecast the external environment, in-cabin intelligence remains strictly recognition-oriented and lacks multi-step rollout capabilities for driver dynamics. We introduce \textbf{Driver-WM}, a driver-centric latent world model that rolls out in-cabin dynamics causally conditioned on out-cabin traffic context. This formulation unifies physical kinematics forecasting with auxiliary behavioral and emotional semantic recognition. Operating in a compact latent space constructed from frozen vision-language features, Driver-WM adopts a dual-stream architecture to separately encode external traffic and internal driver states. These streams are directionally coupled via a gated causal injection mechanism, which uses a learned vector gate to modulate external contextual perturbations while strictly enforcing temporal causality. Experiments on AIDE show robust long-horizon forecasting on reactive high-motion clips, improved driver/traffic semantic alignment, and controlled interventions that expose the external-to-internal mechanism.

\noindent\textbf{Project page:} 
\href{https://fisher75.github.io/haozhuangchi.github.io/driver-wm/}{\texttt{Driver-WM}}

\keywords{World model \and Driver monitor system \and Vision-language model \and Human-in-the-loop driving automation \and Autonomous driving}
\end{abstract}

\begin{figure}[t] 
    \centering
    \includegraphics[width=\textwidth]{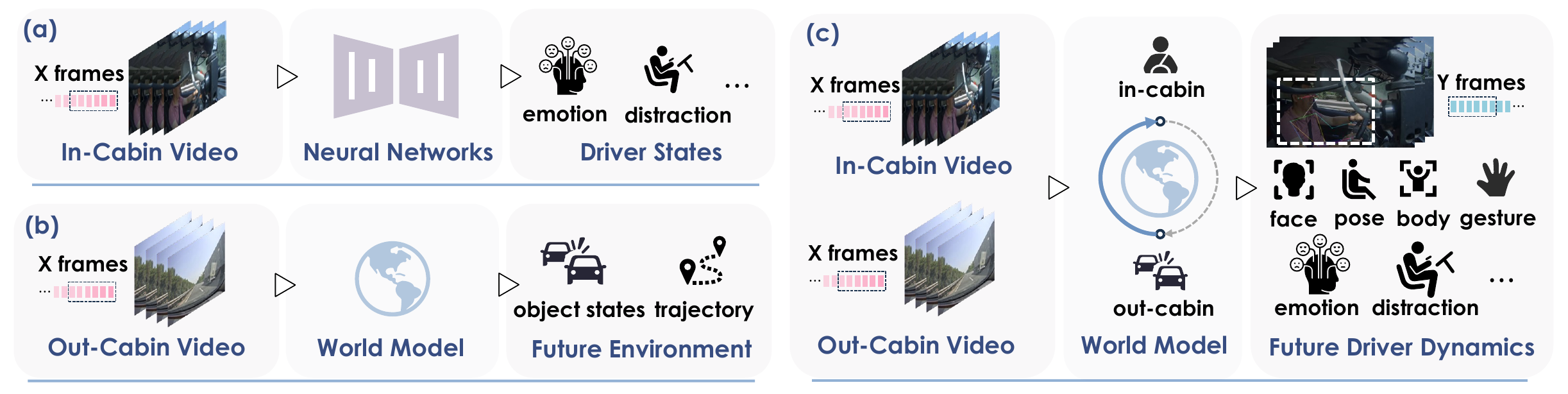}
    \caption{\textbf{The comparison of three paradigms:} (a) Regular driver monitoring systems (DMS) for driver-state recognition. (b) Standard world models for future environment forecasting. (c) \textbf{Driver-WM (ours)} that performs multi-step rollout of internal driver dynamics explicitly conditioned on synchronized external traffic observations.}
    \label{fig:teaser}
\end{figure}

\section{Introduction}
\label{sec:intro}

Autonomous driving has evolved from isolated driver assistance functions toward human-centered intelligent systems~\cite{chen2022milestones,chen2024end, 10844014}. Currently, the majority of real-world driving automation systems are deployed at the SAE L2/L3 level~\cite{1370861704794099599}, where driving responsibility continuously shifts between the autonomous policy and human supervision in a shared-control (mixed-autonomy) setting~\cite{XING2021103199, SU2024109753}. Although recent advances in end-to-end systems~\cite{hu2023poad,li2024law} and vision-language-action (VLA) paradigms~\cite{driess2023palme,brohan2023rt2,huang2026automot} have notably improved the understanding and reasoning capabilities of perception and planning~\cite{hu2023poad,jiang2023vad,liu2025hybrid,10767278}, the ultimate safety of current system domains still requires the human in the loop. In practice, many safety-critical failures are associated with inadequate takeover readiness rather than incorrect scene understanding~\cite{weaver2022systematic}, and the risk increases when the system operates beyond its functional domain~\cite{2cf333680ce943f0ad359635a0df209f} or under evolving traffic interactions~\cite{wang2022social,11543210}.

Motivated by the need to reason about long-horizon safety and interaction, world models have recently emerged as a principled framework for autonomous driving~\cite{kong20253d}. By predicting how the external environment evolves conditioned on current observations and actions, they have been widely adopted for forward simulation, maneuver reasoning, and policy training, ranging from unified full-stack driving systems to generative traffic simulators and driving foundation models~\cite{hu2023gaia1,wang2024drivedreamer,chen2024vista,huang2026towards,xiong2026unidrive,dong2026unifiedworldmodelsvisual,yan2026causaldrive}. To support efficient long-horizon prediction and scalable closed-loop rollout, latent world models have emerged that operate dynamic prediction in compressed latent space~\cite{li2024law,zheng2025world4drive}. However, existing world models are primarily environment-oriented (Fig.~\ref{fig:teaser}b): they model how roads, surrounding agents, and vehicles evolve, while the driver is usually treated as a post-hoc source of risk. They follow an environment-to-action paradigm and do not model how external driving events drive the driver’s internal evolution, leaving posture, gaze, motor readiness, and reaction behavior outside the rollout loop.

In parallel, in-cabin modules, including DMS~\cite{martin2019driveact,yang2023aide,chi2025vlmdm} (Fig.~\ref{fig:teaser}a) and emerging LLM-based cockpit assistants~\cite{sima2024drivelm}, remain recognition- and interaction-oriented rather than predictive. DMS primarily focuses on recognizing instantaneous states such as distraction or fatigue from short temporal windows~\cite{Liu_2025_CVPR,jain2016brain4cars, wang2026micro}, while LLM-based systems emphasize dialogue, intent understanding, and high-level reasoning~\cite{sima2024drivelm,liu2023llava,liao2026farefastslowagenticrobotic}. Although effective for detection and interaction, both paradigms lack the capability to predict how the driver’s physical and cognitive state will evolve over time in response to changing traffic conditions, treating the driver as an observed entity rather than a dynamical system. As a result, in-cabin intelligence remains limited to recognition-based alerts and cannot support anticipatory safety reasoning or systematic analysis of how traffic context modulates driver responses over a prediction horizon. This is critical for L2/L3 driving automation, where safe design requires anticipating not only environment evolution but also driver responses seconds ahead and under hypothetical interventions. A world model that ignores the driver’s internal dynamics is therefore incomplete for shared-control operation.

To address these issues, we propose \textbf{Driver-WM}, a driver-centric latent world model that rolls out future driver dynamics conditioned on synchronized in-cabin and out-cabin observations. In this context, \emph{dynamics} denotes the temporal evolution of the driver's internal state in a compact latent space. This underlying evolution supports both physically grounded \emph{kinematics} (decoded as future skeleton trajectories) and semantic factors (e.g., driver behavior and emotion, predicted as auxiliary regularizers). Instead of generating pixels, Driver-WM forecasts structured 2D skeleton keypoints of the driver’s pose and motion~\cite{guo2023backtomlp,zhu2023motionbert}, providing an efficient target that is stable for long-horizon prediction and naturally aligned with in-cabin supervision. Driver-WM operates in a compact latent space, where frozen vision–language model (VLM)~\cite{bai2025qwen3vl,li2023blip2} features serve as a perceptual encoder to compress high-dimensional visual inputs, and a lightweight world-model core learns latent dynamics with a past/future rollout protocol initialized by observed latents. It adopts a dual-stream latent architecture to separately represent traffic context and driver state, and couples them through a gated interaction module that controls how external driving events drive internal dynamics. This formulation explicitly models how the environment shapes human reactions, enabling both realistic driver response forecasting and controlled test-time interventions for mechanism analysis.

In summary, our contributions are threefold:
\begin{itemize}
\item \textbf{Driver-centric latent world model for dynamics rollout:} multi-step forecasting of in-cabin driver dynamics, unifying kinematic trajectories and auxiliary semantic factors, causally conditioned on observed traffic context.
\item \textbf{Directionally coupled dual-stream dynamics with gated injection:} explicit external-to-internal coupling that supports controllable conditioning and controlled intervention analysis.
\item \textbf{Foundation-derived state interface with unified decoding:} frozen VLM features serve as a compact perceptual interface, supporting geometric rollout with auxiliary semantic regularization.
\end{itemize}

% ============================================================
% 2. Related Work (Re-ordered for Impact)
% ============================================================
\section{Related Work}
\label{sec:related}

\subsection{World Models in Autonomous Driving}
\label{subsec:related_worldmodels}

Most driving world models formulate future prediction as conditional video generation in pixel space. Examples include GAIA-1~\cite{hu2023gaia1}, DriveDreamer~\cite{wang2024drivedreamer,ge2024drivedreamer2}, Vista~\cite{chen2024vista} for controllable rollouts, DriveDreamer4D~\cite{zhao2025drivedreamer4d} for 4D scene generation, UniDrive-WM, UniWM, and DriveVA~\cite{xiong2026unidrive,dong2026unifiedworldmodelsvisual,liu2026drivevavideoactionmodels} for unified planning and generation, and CausalDrive~\cite{yan2026causaldrive} for interactive causal simulation. While pixel-space rollouts offer high visual fidelity, they often entangle semantics, geometry, and physics with low-level rendering, and their iterative diffusion process suffers from inherent latency and extrapolation instability~\cite{zhao2026resilphaseplugandplayphasemapping}. Recent efforts, such as MAD~\cite{rahimi2026mad}, attempt to explicitly decouple motion and appearance~\cite{chengzhijing2026moca} to improve generation efficiency. However, these approaches remain strictly focused on the external environment; enforcing rigorous structural constraints~\cite{li2026optimizationguideddiffusioninteractivescene}
remains challenging in a pixel-generation formulation.
Latent-state world models address this limitation by predicting and planning in a compact state space. In autonomous driving, LAW~\cite{li2024law} regularizes end-to-end driving via temporal consistency, while World4Drive~\cite{zheng2025world4drive} constructs an intention-aware latent model for complex interactions. Yet, whether employing motion-decoupled video generators or latent-state planners, existing methods predominantly model the evolution of the external scene, leaving the human driver unmodeled in the predictive state.
Driver-WM diverges from this scene-centric paradigm by introducing a driver-centric latent dynamics model causally conditioned on out-cabin traffic context. Instead of raw video synthesis, we utilize skeleton trajectory rollout as a physically grounded target, decoupling human kinematics from visual appearance~\cite{Yan_2026_CVPR} and enabling explicit analysis of how external driving events shape predicted driver behavior.

\subsection{Driver and Cabin State Modeling}
\label{subsec:related_driver}

In-cabin perception is commonly studied as discrete state recognition, aiming to infer discrete driver states or activities, such as distraction or drowsiness~\cite{yang2023quantitative,xing2019driver}, from visual observations~\cite{chi2025vlmdm,wang2026micro}. Drive\&Act~\cite{martin2019driveact} provides a benchmark for fine-grained driver activity recognition with multi-modal streams, and AIDE~\cite{yang2023aide} further links in-cabin observations with synchronized out-cabin traffic context. Recent unified multi-task frameworks have reported strong semantic recognition results on this benchmark; for example, MMTL-UniAD~\cite{Liu_2025_CVPR}. Driver-related prediction has also been explored in maneuver/intention anticipation, e.g., Brain4Cars~\cite{jain2016brain4cars}. These works primarily focus on predicting discrete categorical labels. As such, they do not explicitly model continuous driver motion responses to external driving events (e.g., ego maneuvers such as braking or turning), which are important for predictive safety assessment.
Human motion forecasting predicts future skeletal trajectories from historical motion. SiMLPe~\cite{guo2023backtomlp} shows that compact architectures can capture strong kinematic priors, while MotionBERT~\cite{zhu2023motionbert} demonstrates the benefit of transferable skeleton representations via large-scale pretraining. Recent studies further incorporate context and interactions: TRiPOD~\cite{adeli2021tripod} studies pose forecasting in the wild, and Waymo-3DSkelMo~\cite{zhu2025waymo3dskelmo} models pedestrian skeletal reactions under traffic interactions. Additionally, explicitly predicting human kinematics, such as hand trajectories, has recently proven highly effective for conditioning future visual and action forecasting in egocentric scenarios~\cite{zhang2025egocentricpredictivemodelconditioned}. In contrast, driver motion inside the cabin is still often treated as isolated estimation without explicit conditioning on the evolving traffic context.
To bridge the gap between classification-based driver monitoring and motion forecasting without explicit conditioning on out-cabin traffic context,
Driver-WM explicitly parameterizes the directional coupling from out-of-cabin driving events to driver motion, formulating traffic-conditioned driver responses as latent dynamics and decoding them into continuous skeleton rollouts.

\subsection{Foundation Models for Embodied Perception}
\label{subsec:relatedF_foundation}

Vision--language foundation models (VLMs) provide open-vocabulary and visual-knowledge representations~\cite{jiang2025vknowuevaluatingvisualknowledge} and are widely used as perception backbones; instruction-tuned families such as Qwen-VL~\cite{bai2024qwenvl, wang2024qwen2vl, bai2025qwen3vl} further improve semantic grounding. Many works adopt a frozen-backbone paradigm~\cite{li2023blip2}, keeping large encoders fixed while learning lightweight interfaces for downstream adaptation.
In embodied AI and autonomous driving, VLM/LLM-based agents are frequently used for high-level reasoning and planning~\cite{NEURIPS2025_087f7678,lin2026models}, including general embodied systems such as PaLM-E~\cite{driess2023palme}, RT-2~\cite{brohan2023rt2}, and FARE~\cite{liao2026farefastslowagenticrobotic}, language-conditioned navigation models~\cite{dong2026languageconditionedworldmodelingvisual}, as well as driving-oriented VLMs like DriveLM~\cite{sima2024drivelm}. Rather than utilizing VLMs to produce discrete symbolic outputs, Driver-WM treats frozen Qwen3-VL features as a continuous state interface for latent dynamics, ensuring that the rollout relies on grounded visual perception rather than language priors~\cite{wu2026lavitaligninglatentvisual,zhu2026medsynapsevbridgingvisualperception}. This formulation enables efficient multi-step physical rollouts that remain semantically consistent with the visual scene, bypassing the need for pixel-level generation.
% ============================================================
% 3. Method (The Core)
% ============================================================
\section{Method}
\label{sec:method}

\subsection{Problem Formulation}
\label{subsec:overview}

We formulate driver-centric cabin world modeling as externally conditioned temporal forecasting. We use in-/out-cabin to denote temporally synchronized raw observations, and internal/external to denote their derived latent states. Given these streams, we learn a world model to roll out the driver's internal dynamics conditioned on the observed external context, and decode the resulting states into physically plausible skeleton trajectories.
In contrast to conventional driver monitoring that focuses on per-frame recognition, our primary objective is multi-step rollout; semantic predictions are used only as auxiliary regularizers.
Throughout this paper, \emph{dynamics} refers to the latent-state evolution governing the driver's future responses, whereas \emph{kinematics} refers specifically to the decoded geometric skeleton trajectories. Semantic predictions provide clip-level regularization and are not used as temporal conditioning inputs.

Let $\mathbf{o}^{\text{in}}_{1:T}=\{\mathbf{o}^{\text{in}}_t\}_{t=1}^{T}$ and $\mathbf{o}^{\text{out}}_{1:T}=\{\mathbf{o}^{\text{out}}_t\}_{t=1}^{T}$ denote synchronized in-cabin and out-cabin observations over horizon $T$.
Driver-WM treats $\mathbf{o}^{\text{out}}$ as an external context that conditions the driver's internal evolution.

Driver-WM adopts a past/future protocol with $T=T_{\mathrm{obs}}+T_{\mathrm{pred}}$ with causal rollouts.
Both streams are initialized with ground-truth latents for $t\le T_{\mathrm{obs}}$.
For $t>T_{\mathrm{obs}}$, Driver-WM performs closed-loop rollouts of external and internal latents, while enforcing that the update at step $t{+}1$ only accesses histories up to step $t$. Crucially, the external rollout serves exclusively to sustain continuous environmental conditioning for the driver, rather than to reconstruct future traffic scenes explicitly.

Dual-stream latent states are formulated in Driver-WM. Specifically, we represent the driver state at time $t$ as an internal latent $\mathbf{z}^{\text{int}}_t \in \mathbb{R}^{D}$ and a set of multi-view external latents $\mathbf{Z}^{\text{ext}}_{t}=\{\mathbf{z}^{\text{ext}}_{t,v}\in\mathbb{R}^{D}\}_{v=1}^{V}$, where $V$ is the number of out-cabin views (e.g., front/left/right) and $D{=}2048$ is the hidden dimension of the frozen Qwen3-VL vision encoder.
Accordingly, the external history is $\mathbf{Z}^{\text{ext}}_{\le t}=\{\mathbf{z}^{\text{ext}}_{\tau,v}\}_{\tau\le t,\,v}$. We further define a pooled external vector $\bar{\mathbf{z}}^{\text{ext}}_{t}=\frac{1}{V}\sum_{v=1}^{V}\mathbf{z}^{\text{ext}}_{t,v}$ and its history $\bar{\mathbf{Z}}^{\text{ext}}_{\le t}=\{\bar{\mathbf{z}}^{\text{ext}}_{\tau}\}_{\tau\le t}$, which is used by the dynamics core.

During the rollout stage, the external context of Driver-WM is advanced by
$\hat{\bar{\mathbf{z}}}^{\text{ext}}_{t+1}=f^{\text{ext}}_{\theta}(\hat{\bar{\mathbf{z}}}^{\text{ext}}_{t})$,
and the internal state is updated by an externally conditioned transition:
\begin{equation}
\label{eq:rollout_core}
\hat{\mathbf{z}}^{\text{int}}_{t+1}
=
\mathcal{F}_{\theta}\!\left(\hat{\mathbf{Z}}^{\text{int}}_{\le t},\; \hat{\bar{\mathbf{Z}}}^{\text{ext}}_{\le t}\right).
\end{equation}
We initialize $\hat{\mathbf{z}}^{\text{int}}_{1:T_{\mathrm{obs}}}=\mathbf{z}^{\text{int}}_{1:T_{\mathrm{obs}}}$ and
$\hat{\bar{\mathbf{z}}}^{\text{ext}}_{1:T_{\mathrm{obs}}}=\bar{\mathbf{z}}^{\text{ext}}_{1:T_{\mathrm{obs}}}$, and compute rollout losses only on the future window.
Driver-WM formulates multi-task decoding with auxiliary regularizers.
A geometric head decodes rolled-out internal latents into skeleton keypoints:
\begin{equation}
\label{eq:skel_decode}
\hat{\mathbf{s}}_{t}=D_{\text{skel}}\!\left(\hat{\mathbf{z}}^{\text{int}}_{t}\right),
\quad \hat{\mathbf{s}}_{t}\in\mathbb{R}^{K\times 2}\;(\text{or }\mathbb{R}^{K\times 3}),
\end{equation}
where $K$ is the number of keypoints (e.g., $K{=}136$ for HALPE-style skeleton).
We also attach lightweight auxiliary heads on $\mathbf{z}^{\text{int}}$ and $\mathbf{z}^{\text{ext}}$ to regularize the latent space with semantic factors.

\begin{figure*}[t]
    \centering
    \includegraphics[width=\textwidth]{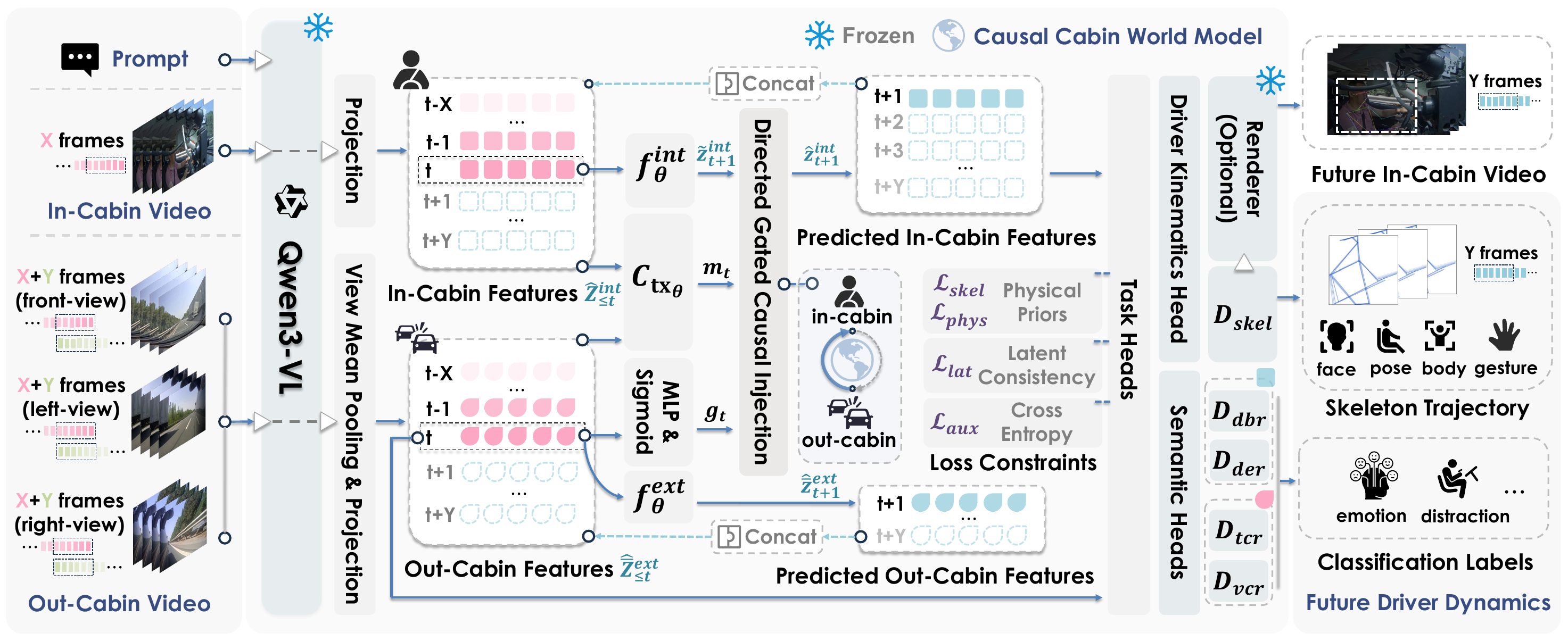} 
    \caption{
    \textbf{Overall Architecture of Driver-WM.} 
    From synchronized in/out-cabin videos, a frozen Qwen3-VL extracts dual-stream latent features. 
    Pooled external history $\hat{\bar{\mathbf{Z}}}^{\text{ext}}_{\le t}$ perturbs the internal transition via a directed Gated Causal Injection with a vector gate $\mathbf{g}_t$, yielding an updated internal latent $\hat{\mathbf{z}}^{\text{int}}_{t+1}$.
    Internal latents are autoregressively rolled out to forecast future states, decoded into skeleton trajectories and auxiliary semantic predictions (e.g., driver behavior) for regularization.
    }
    \label{fig:main_architecture}
\end{figure*}

\subsection{VLM-based Perception and Latent Interface}
\label{subsec:perception}

We model temporal dynamics in a compact semantic space rather than raw pixels.
We adopt \textbf{Qwen3-VL}~\cite{bai2025qwen3vl} as a frozen perceptual backbone and use its representation as the latent interface, following fixed-encoder visual-interface adaptation paradigms (e.g., BLIP-2~\cite{li2023blip2} and training-free VLM adaptation~\cite{Debnath_2026_CVPR}).

\noindent\textbf{Frozen features.}
Let $E_{\text{vlm}}(\cdot)$ denote the visual perception stack of Qwen3-VL.
Given synchronized observations at time $t$, we extract
$\mathbf{f}^{\text{in}}_{t}=E_{\text{vlm}}(\mathbf{o}^{\text{in}}_{t})$ and
$\mathbf{f}^{\text{out}}_{t,v}=E_{\text{vlm}}(\mathbf{o}^{\text{out}}_{t,v})$,
where $v\in\{1,\dots,V\}$ indexes the out-cabin views and $D{=}2048$.
We pre-extract and cache $\{\mathbf{f}^{\text{in}}_{t},\mathbf{f}^{\text{out}}_{t}\}_{t=1}^{T}$ and exclude the VLM from the training graph to reduce computation and stabilize optimization, so that learning focuses on the world-model core for driver dynamics.

\noindent\textbf{View-conditioned interface.}
To disambiguate camera sources in a shared semantic space, we inject a learnable view embedding
\begin{equation}
\bar{\mathbf{f}}^{\text{in}}_{t}=\mathbf{f}^{\text{in}}_{t}+\mathbf{e}_{\text{view}}(\text{in}),\qquad
\bar{\mathbf{f}}^{\text{out}}_{t,v}=\mathbf{f}^{\text{out}}_{t,v}+\mathbf{e}_{\text{view}}(v),\qquad
\mathbf{e}_{\text{view}}(\cdot)\in\mathbb{R}^{D},
\end{equation}
where $v$ is the view identifier (e.g., $\{\text{front},\text{left},\text{right},\text{in-cabin}\}$). We obtain $\bar{\mathbf{f}}^{\text{in}}_{t}$ and $\bar{\mathbf{f}}^{\text{out}}_{t,v}$ accordingly.

\noindent\textbf{Latent State Interface.}
By default, we use an identity latent interface:
\begin{equation}
\mathbf{z}^{\text{int}}_{t}=\bar{\mathbf{f}}^{\text{in}}_{t},\qquad
\mathbf{z}^{\text{ext}}_{t,v}=\bar{\mathbf{f}}^{\text{out}}_{t,v}.
\end{equation}
We apply mean pooling across views to obtain $\bar{\mathbf{z}}^{\text{ext}}_{t}=\frac{1}{V}\sum_{v=1}^{V}\mathbf{z}^{\text{ext}}_{t,v}$.

\subsection{Causal Driver World Model}
\label{subsec:core}

We formulate the world model as a lightweight gated module, in contrast to heavy block-based architectures. Given dual-stream latents $\{(\mathbf{z}^{\text{int}}_{t},\bar{\mathbf{z}}^{\text{ext}}_{t})\}_{t=1}^{T}$ (Sec.~\ref{subsec:overview}--\ref{subsec:perception}), we learn a \emph{directed} world model that rolls out the driver's internal dynamics conditioned on external context. We enforce two inductive biases: \textbf{(i) temporal causality} (the update at $t{+}1$ only depends on context up to $t$) and \textbf{(ii) directed coupling} (external $\rightarrow$ internal via an explicit gate rather than symmetric fusion). In our main configuration, we apply a causal temporal self-attention pre-encoding to each stream before rollout; the bidirectional variant is used only as a non-causal upper bound.

\noindent\textbf{Causal context summary.}
Let $\hat{\mathbf{Z}}^{\text{int}}_{\le t}$ and $\hat{\bar{\mathbf{Z}}}^{\text{ext}}_{\le t}$ denote the internal/external histories available up to step $t$.
We compute an external-to-internal context summary $\mathbf{m}_t\in\mathbb{R}^{D}$ via cross-attention:
\begin{equation}
\label{eq:ctx_attn}
\mathbf{m}_{t} = \mathrm{Ctx}_{\theta}\!\left(\hat{\mathbf{Z}}^{\text{int}}_{\le t},\, \hat{\bar{\mathbf{Z}}}^{\text{ext}}_{\le t}\right).
\end{equation}
Temporal causality is enforced by truncating the histories passed to $\mathrm{Ctx}_{\theta}$ at each rollout step (i.e., only providing representations $\le t$). 

\noindent\textbf{Internal transition.}
We predict a candidate next internal latent via a lightweight transition predictor $f^{\text{int}}_{\theta}$:
\begin{equation}
\label{eq:int_transition}
\tilde{\mathbf{z}}^{\text{int}}_{t+1}
= f^{\text{int}}_{\theta}\!\left(\hat{\mathbf{z}}^{\text{int}}_{t}\right).
\end{equation}
In practice, we parameterize a diagonal Gaussian transition during training and use the mean prediction for deterministic inference; KL regularization is enabled only in the probabilistic ablation (Supplementary).

\noindent\textbf{Gated causal coupling.}
External driving events do not affect the driver equally; we introduce a vector-valued gate $\mathbf{g}_t\in(0,1)^D$ to regulate contextual perturbation:
\begin{equation}
\label{eq:gated_fusion}
\mathbf{g}_t=\sigma\!\left(\mathrm{MLP}_{g}\!\left(\hat{\bar{\mathbf{z}}}^{\text{ext}}_{t}\right)\right),
\qquad
\hat{\mathbf{z}}^{\text{int}}_{t+1} 
= (\mathbf{1}-\mathbf{g}_t)\odot \tilde{\mathbf{z}}^{\text{int}}_{t+1}
\;+\;
\mathbf{g}_t\odot \mathbf{m}_{t}.
\end{equation}
For controlled interventions analysis, we bypass the learned gate with a scalar override $\lambda_{\mathrm{CA}}$,
$\hat{\mathbf{z}}^{\text{int}}_{t+1}=(1-\lambda_{\mathrm{CA}})\tilde{\mathbf{z}}^{\text{int}}_{t+1}+\lambda_{\mathrm{CA}}\mathbf{m}_t$.

\subsection{Unified Decoding and Optimization Objectives}
\label{subsec:decode_loss}

Driver-WM rolls out internal latents $\hat{\mathbf{z}}^{\text{int}}_{1:T}$ and uses them as a shared substrate for (i) skeleton rollout (primary) and (ii) auxiliary semantic predictions (regularizers). We adopt a lightweight \emph{one core, multiple heads} design to keep the temporal core compact.

\noindent\textbf{Geometric decoding.}
We decode the rolled-out internal latents into skeleton keypoints via a Spatial-Temporal Graph Convolutional Network (ST-GCN):
\begin{equation}
\label{eq:skel_head}
\hat{\mathbf{s}}_{t}=D_{\text{skel}}\!\left(\hat{\mathbf{z}}^{\text{int}}_{t}\right).
\end{equation}
For 2D rollout, $\hat{\mathbf{s}}_{t}\in[0,1]^{K\times 2}$ stores normalized joint coordinates (sigmoid output), where $K$ is the number of joints (e.g., $K{=}136$ for HALPE). $D_{\text{skel}}$ uses human skeletal adjacency as the graph topology. We supervise $\{\hat{\mathbf{s}}_t\}$ with regression loss $\mathcal{L}_{\text{skel}}$ over the rollout horizon; model selection is driven by multi-step skeleton accuracy (e.g., MPJPE). These are image-plane quantities; ROI-/vehicle-frame use requires calibrated cabin references and camera calibration, while metric 3D can use priors, monocular lifting, or depth/multi-view sensors.

\noindent\textbf{Physical priors for plausible rollouts.}
Beyond pointwise regression, we regularize decoded trajectories with
\begin{equation}
\label{eq:total_phys_loss}
\mathcal{L}_{\text{phys}}
=
\lambda_{\text{bone}}\mathcal{L}_{\text{bone}}
+
\lambda_{\text{smooth}}\mathcal{L}_{\text{smooth}}
+
\lambda_{\text{seat}}\mathcal{L}_{\text{seat}},
\end{equation}
where $\mathcal{L}_{\text{bone}}$ preserves kinematic edge lengths, $\mathcal{L}_{\text{smooth}}$ suppresses temporal jitter, and $\mathcal{L}_{\text{seat}}$ is an optional cabin/seat feasibility term.
For the kinematic tree $\mathcal{E}$, we use
\begin{equation}
\label{eq:bone_loss}
\mathcal{L}_{\text{bone}}
=
\sum_{t}\sum_{(i,j)\in\mathcal{E}}
\left|
\left\|\hat{\mathbf{s}}_{t,i}-\hat{\mathbf{s}}_{t,j}\right\|_2
-
\left\|\mathbf{s}_{t,i}-\mathbf{s}_{t,j}\right\|_2
\right|.
\end{equation}
Temporal smoothness is enforced by matching first-order motion and penalizing second-order jitter:
\begin{equation}
\label{eq:smooth_loss}
\mathcal{L}_{\text{smooth}}
=
\sum_{t}\left\|\Delta \hat{\mathbf{s}}_{t}-\Delta \mathbf{s}_{t}\right\|_2^2
+
\sum_{t}\left\|\Delta^2 \hat{\mathbf{s}}_{t}\right\|_2^2,
\end{equation}
where $\Delta \hat{\mathbf{s}}_{t}=\hat{\mathbf{s}}_{t+1}-\hat{\mathbf{s}}_{t}$ and
$\Delta^2 \hat{\mathbf{s}}_{t}=\hat{\mathbf{s}}_{t+2}-2\hat{\mathbf{s}}_{t+1}+\hat{\mathbf{s}}_{t}$.
The 3D counterpart and implementation details of $\mathcal{L}_{\text{seat}}$ are provided in the Supplementary Material.

\noindent\textbf{Auxiliary semantic heads.}
We apply last-step pooling for clip-level recognition:
\begin{align}
\label{eq:int_aux}
\hat{\mathbf{y}}^{\text{dbr}}&=D_{\text{dbr}}(\hat{\mathbf{z}}^{\text{int}}_{T}),\qquad
\hat{\mathbf{y}}^{\text{der}}=D_{\text{der}}(\hat{\mathbf{z}}^{\text{int}}_{T}),\\
\label{eq:ext_aux}
\hat{\mathbf{y}}^{\text{tcr}}&=D_{\text{tcr}}(\bar{\mathbf{z}}^{\text{ext}}_{T}),\qquad
\hat{\mathbf{y}}^{\text{vcr}}=D_{\text{vcr}}(\bar{\mathbf{z}}^{\text{ext}}_{T}),
\end{align}
trained with cross-entropy losses and used exclusively as auxiliary regularizers. External semantics are decoded from the final external context $\bar{\mathbf{z}}^{\text{ext}}_{T}$, while internal semantics use the rolled-out state $\hat{\mathbf{z}}^{\text{int}}_{T}$.

\noindent\textbf{Latent rollout consistency.}
Besides decoded skeleton supervision, we regularize future internal rollouts by either direct latent matching or velocity matching:
\begin{equation}
\label{eq:latent_losses}
\mathcal{L}_{\text{lat}}^{\text{direct}}
=\sum_{t}\|\hat{\mathbf{z}}^{\text{int}}_{t+1}-\mathbf{z}^{\text{int}}_{t+1}\|_2^2,\quad
\mathcal{L}_{\text{lat}}^{\text{vel}}
=\sum_{t}\|(\hat{\mathbf{z}}^{\text{int}}_{t+1}-\hat{\mathbf{z}}^{\text{int}}_{t})-(\mathbf{z}^{\text{int}}_{t+1}-\mathbf{z}^{\text{int}}_{t})\|_2^2.
\end{equation}

The final objective aggregates latent consistency, skeleton supervision, physical priors, auxiliary semantics:
\begin{equation}
\label{eq:total_loss}
\mathcal{L}
=
\lambda_{\text{lat}}\mathcal{L}_{\text{lat}}
+
\lambda_{\text{skel}}\mathcal{L}_{\text{skel}}
+
\lambda_{\text{aux}}\mathcal{L}_{\text{aux}}
+
\lambda_{\text{phys}}\mathcal{L}_{\text{phys}},
\end{equation}

% ============================================================
% 4. Experiments
% ============================================================
\section{Experiments}
\label{sec:exp}

\subsection{Experimental Setup}
\label{subsec:exp_setup}

\noindent\textbf{Dataset, protocol, and targets.}
We evaluate Driver-WM on the AIDE assistive driving benchmark~\cite{yang2023aide} to validate its capacity for both kinematic rollout and auxiliary semantic recognition. Each video is segmented into 3-second clips with 10 uniformly sampled frames ($\sim$3.3\,Hz). We follow a $5{\rightarrow}5$ time-causal protocol: given $T_{\mathrm{obs}}{=}5$ observed steps, we roll out $T_{\mathrm{pred}}{=}5$ future steps in latent space, decoding the internal states into HALPE-136 2D skeletons ($K{=}136$). Rollout losses (latent and skeleton) are computed exclusively on the future window ($t > T_{\mathrm{obs}}$). We attach four auxiliary classification heads as semantic regularizers: 7-class Driver Behavior (DBR) and 5-class Emotion (DER) for the in-cabin stream, alongside 3-class Traffic Context (TCR) and 5-class Vehicle Condition (VCR) for the out-cabin stream. Ground-truth semantic labels are used exclusively to supervise these auxiliary heads. They are strictly excluded from serving as conditioning inputs to the temporal transition or gating pathways (i.e., label embeddings are disabled) to ensure a purely vision-driven rollout.

\noindent\textbf{Implementation details.}
We pre-extract per-frame features ($D{=}2048$) using a frozen Qwen3-VL-2B~\cite{bai2025qwen3vl} with a fixed prompt to avoid label leakage. Models are trained for 80 epochs using AdamW (learning rate $5{\times}10^{-5}$, batch size 32) on A5000 GPUs. In inference, temporal causality is enforced by truncating the internal/external histories passed to the context interaction and gating modules at each step (strictly $\le t$). Unless stated otherwise, models are trained from scratch; warm-start variants are detailed in the Supplementary. We additionally verify strict zero-lookahead numerically: replacing the unobservable future external suffix with noise or zeros changes the predicted skeletal tensors by at most $5\times10^{-7}$ (max abs diff; see Supplementary).

\noindent\textbf{Metrics and baselines.}
We report horizon-averaged MPJPE (px), diagonal-normalized MPJPE (d-nMPJPE, \%), and PCK@0.05; PCK@0.10 and per-head Accuracy are deferred to the Supplementary.
We compute d-nMPJPE as $100\times\mathrm{MPJPE}/d$ with the image diagonal $d=\sqrt{1920^2+1080^2}$, and use a threshold of $0.05d$ for PCK@0.05.
Semantic performance is measured by Macro-F1.
We compare against motion-only forecasters (Zero-Velocity~\cite{Martinez_2017_CVPR}, ST-GCN~\cite{yan2018spatial}, SiMLPe~\cite{guo2023backtomlp}, MotionBERT~\cite{zhu2023motionbert}) and an offline encoder-decoder reference~\cite{Martinez_2017_CVPR} representing the non-causal motion bound.
To rigorously isolate architectural gains from visual backbone disparities, we additionally construct controlled contextual baselines (e.g., static pooling, single-stream, late fusion) under the exact same frozen Qwen3-VL interface. All learnable methods follow the identical $5{\rightarrow}5$ rollout protocol.

\subsection{Main Results}
\label{subsec:main_results}

\begin{table*}[t]
\centering
\caption{\textbf{Main results on AIDE under the $5{\rightarrow}5$ rollout.}
We report horizon-averaged geometric errors (MPJPE in px, d-nMPJPE in \%) and semantic Macro-F1.
}
\label{tab:main_results}

\small
\setlength{\tabcolsep}{3.5pt}
\resizebox{\textwidth}{!}{%
\begin{tabular}{l cc cc ccccc}
\toprule
\textbf{Model} &
\multicolumn{2}{c}{\textbf{All}} &
\multicolumn{2}{c}{\textbf{HM}} &
\textbf{PCK} & \textbf{DBR} & \textbf{DER} & \textbf{TCR} & \textbf{VCR} \\
\cmidrule(lr){2-3} \cmidrule(lr){4-5}
& \textbf{MPJPE}$_{\downarrow}$ & \textbf{d-nMPJPE}$_{\downarrow}$
& \textbf{MPJPE}$_{\downarrow}$ & \textbf{d-nMPJPE}$_{\downarrow}$
& \textbf{@0.05}$_{\uparrow}$ & \textbf{F1}$_{\uparrow}$ & \textbf{F1}$_{\uparrow}$ & \textbf{F1}$_{\uparrow}$ & \textbf{F1}$_{\uparrow}$ \\
\midrule
\multicolumn{10}{l}{\textit{Motion-only \& offline baselines}} \\
\textcolor{dgray}{Zero-Velocity}$\dagger$~\cite{Martinez_2017_CVPR} & \textcolor{dgray}{52.89} & \textcolor{dgray}{2.40} & \textcolor{dgray}{139.19} & \textcolor{dgray}{6.32} & \textcolor{dgray}{85.95} & \textcolor{dgray}{8.27} & \textcolor{dgray}{14.94} & -- & -- \\
ST-GCN~\cite{yan2018spatial}      & 110.98 & 5.04 & 158.10 & 7.18 & 60.97 & 18.96 & 22.66 & -- & -- \\
SiMLPe~\cite{guo2023backtomlp}        & 106.38 & 4.83 & 156.29 & 7.09 & 63.45 & 37.26 & 41.69 & -- & -- \\
MotionBERT~\cite{zhu2023motionbert}   & 73.51 & 3.34 & 141.53 & 6.42 & 78.01 & 56.70 & 52.71 & -- & -- \\
Offline Enc-Dec~\cite{Martinez_2017_CVPR} & 75.87 & 3.45 & 144.05 & 6.54 & 77.17 & 53.82 & 50.68 & -- & -- \\
\midrule
\multicolumn{10}{l}{\textit{Controlled contextual baselines (identical VLM interface)}} \\
Static pooling (no rollout)$\ddagger$  & \textbf{68.50} & \textbf{3.11} & \textbf{134.56} & \textbf{6.11} & \textbf{72.55} & 54.04 & 60.07 & 26.39 & 14.00 \\
Single-stream & 90.87 & 4.13 & 147.44 & 6.69 & 59.63 & 24.56  & 24.35 & 26.39 & 14.00 \\
Late fusion   & 82.59 & 3.75 & 143.31 & 6.51 & 65.07 & 45.87 & 49.10 & 87.85 & 65.97 \\
Cross-Attn only & 80.14 & 3.64 & 142.41 & 6.47 & 66.43 & 62.52 & 68.75 & 87.94 & \textbf{69.09} \\
\midrule
\multicolumn{10}{l}{\textit{Ours}} \\
\rowcolor{gray!10}
\textbf{Driver-WM (main)} & \textbf{71.47} & \textbf{3.24} & 138.03 & 6.27 & \textbf{71.66} & 68.07 & 72.61 & 90.15 & 68.34 \\
Driver-WM (full pretrained) & 72.68 & 3.30 & 138.03 & 6.27 & 70.99 & 73.22 & 73.97 & \textbf{90.54} & 62.09 \\
Non-causal (bidir) & 72.15 & 3.27 & \textbf{136.50} & \textbf{6.20} & 71.39 & \textbf{73.35} & \textbf{74.46} & 88.65 & \textbf{68.82} \\
\bottomrule
\end{tabular}%
}

\begin{minipage}{\textwidth}
\tiny
$^{\dagger}$ Copy-last-frame baseline favored by $\ell_2$ metrics under inertial continuity.\\
$^{\ddagger}$ Static pooling baseline (no rollout) can yield low horizon-averaged $\ell_2$ errors via mean-regression. \\
$-$  Externally grounded semantics (TCR/VCR) are inapplicable for motion-only baselines.\\
\textbf{HM} The High-Motion subset: the Top 10\% clips ranked by a ground-truth motion score computed on the future window as the mean per-joint frame-to-frame displacement in pixel coordinates (contains 60/609 test clips; see Supplementary).
\end{minipage}
\end{table*}

Table~\ref{tab:main_results} reports the comparison under the fixed $5{\rightarrow}5$ causal rollout. To focus on paradigm-level differences, we defer dedicated ablations and controlled variants to Sec.~\ref{subsec:ablation}.

\noindent\textbf{Motion-only baselines and an inertia effect under $\ell_2$-style metrics.}
An apparent outlier in Table~\ref{tab:main_results} is the Zero-Velocity baseline, which attains the lowest overall d-nMPJPE. This is consistent with naturalistic driving clips dominated by mild motion and inertial continuity: copying the last observed pose exploits this inertia to reduce an $\ell_2$-style geometric metric globally. However, autonomous safety hinges on reactive, high-dynamic moments. As illustrated in Fig.~\ref{fig:cf_qual}(b), when focusing on the safety-critical High-Motion tail, the Zero-Velocity baseline's error grows sharply as the horizon extends. While simpler baselines benefit from strong spatial priors at initial steps, Driver-WM successfully captures reactive dynamics, significantly mitigating the rapid error accumulation observed in motion-only models and achieving competitive long-horizon fidelity (e.g., $h{=}5$ MPJPE 155.82 vs. S0 178.62) on the High-Motion tail. While stronger motion-only forecasters (e.g., MotionBERT) improve kinematic fidelity, they do not explicitly condition on out-cabin context and degrade on the High-Motion tail at longer horizons. Finally, the offline variant serves strictly as a non-causal reference.

\noindent\textbf{Contextual paradigms and a static mean-regression trap.}
Introducing external visual context changes the trade-off substantially under the same frozen Qwen3-VL interface.
Notably, the static pooling baseline (no rollout) can appear strong on horizon-averaged $\ell_2$ metrics, including on the HM subset, since a time-invariant, time-averaged hypothesis may reduce the average penalty when reactive maneuvers exhibit temporal ambiguity under sparse sampling.
However, this baseline does not produce an autoregressive trajectory and therefore cannot support horizon-wise degradation analysis or test-time mechanism probes.
In addition, its externally grounded Macro-F1 remains low (Table~\ref{tab:main_results}), indicating limited balanced recognition of traffic context.
In contrast, Driver-WM explicitly models multi-step dynamics conditioned on traffic observations and yields robust long-horizon rollouts on the safety-critical High-Motion tail (Fig.~\ref{fig:cf_qual}(b)).

\noindent\textbf{Driver-WM.}
Driver-WM maintains competitive horizon-averaged errors while improving long-horizon behavior on the HM tail (Fig.~\ref{fig:cf_qual}b). The static pooling baseline can reduce the averaged $\ell_2$ error by mean regression, but produces no autoregressive trajectory and cannot support mechanism probes. Driver-WM instead performs time-causal rollouts conditioned on traffic observations, improving semantic consistency under the same frozen interface. The non-causal variant is an offline reference and does not yield consistent gains, supporting causal modeling as a deployment-consistent bias.

\noindent\textbf{Robustness.}
Across three random initialization seeds, the main Driver-WM configuration demonstrates high stability, yielding a consistent All d-nMPJPE of $3.25{\pm}0.01\%$ alongside a robust TCR F1 score of $89.26{\pm}1.33$. In contrast, sub-optimal fusion designs exhibit notably larger variance; for instance, the M2 variant fluctuates with an All MPJPE of $79.93{\pm}2.35$\,px. The complete multi-seed table and a reference comparison to recent classification-only methods are provided in the Supplementary Material. Note that these approaches do not perform kinematic rollout and are therefore not directly comparable to our setting.

\noindent\textbf{Risk-ranking signal.}
Using only Driver-WM outputs, a composite of predicted motion, DBR/DER probabilities, and uncertainty ranks GT-HM events better than motion-only: AUPRC 0.1785 vs. 0.1364, with random $\sim$0.10 and paired-bootstrap CI $[0.013,0.094]$ ($p{=}0.0009$). This suggests that structured driver rollouts provide safety-relevant risk cues.

\subsection{Ablation Studies on Driver-WM Architecture}
\label{subsec:ablation}

To isolate the contributions of key design choices in Driver-WM, we evaluate controlled variants around the main model under the identical $5{\rightarrow}5$ causal rollout. Table~\ref{tab:ablation} summarizes representative ablations.

\begin{table}[t]
\centering
\caption{\textbf{Ablations on Driver-WM.} Protocol and metrics follow Table~\ref{tab:main_results}. Geometric metrics are inapplicable for the \textit{no pose head} variant.}
\label{tab:ablation}
\small
\setlength{\tabcolsep}{3.2pt}
\resizebox{\columnwidth}{!}{%
\begin{tabular}{l cc cc c cccc}
\toprule
\textbf{Variant} &
\multicolumn{2}{c}{\textbf{All}} &
\multicolumn{2}{c}{\textbf{HM}} &
\textbf{PCK} &
\textbf{DBR} &
\textbf{DER} &
\textbf{TCR} &
\textbf{VCR} \\
\cmidrule(lr){2-3} \cmidrule(lr){4-5}
& \textbf{MPJPE}$_{\downarrow}$ & \textbf{d-nMPJPE}$_{\downarrow}$ 
& \textbf{MPJPE}$_{\downarrow}$ & \textbf{d-nMPJPE}$_{\downarrow}$
& \textbf{@0.05}$_{\uparrow}$
& \textbf{F1}$_{\uparrow}$ & \textbf{F1}$_{\uparrow}$ & \textbf{F1}$_{\uparrow}$ & \textbf{F1}$_{\uparrow}$ \\
\midrule
\rowcolor{gray!10}
Driver-WM (main) & 71.47 & 3.24 & 138.03 & 6.27 & 71.66 & 68.07 & 72.61 & 90.15 & 68.34 \\
\midrule
\multicolumn{10}{l}{\textit{External Context}} \\
w/o ext context & 71.74 & 3.26 & 136.94 & 6.21 & 71.13 & 71.33 & 63.56 & 26.39 & 14.00 \\
\midrule
\multicolumn{10}{l}{\textit{Dynamics Core \& Training Choices}} \\
RSSM/GRU dyn.   & 78.42 & 3.56 & 136.46 & 6.19 & 67.48 & 67.15 & 72.29 & 88.26 & 63.74 \\
+Skeleton-Pre   & 70.92 & 3.22 & 136.04 & 6.18 & 72.09 & 69.55 & 75.65 & 88.33 & 70.24 \\
\midrule
\multicolumn{10}{l}{\textit{Regularization \& Interface Diagnostics}} \\
KL bottleneck   & 70.56 & 3.20 & 136.50 & 6.20 & 71.71 & 69.89 & 71.54 & 88.84 & 64.68 \\
no-prompt feat. & 76.54 & 3.47 & 142.08 & 6.45 & 68.28 & 68.88 & 69.92 & 76.87 & 58.59 \\
no pose head    & --    & --   & --     & --   & --    & 67.22 & 70.92 & 88.83 & 69.44 \\
\bottomrule
\end{tabular}%
}
\end{table}

\noindent\textbf{External context is necessary for externally grounded semantics.}
Severing the out-cabin stream leads to a substantial degradation on the externally grounded head (TCR), while also reducing DER. Although re-training without external context yields competitive averaged MPJPE due to dominant low-motion segments, it lacks access to environmental triggers. This gap is revealed by test-time interventions on the jointly trained model: removing external context increases rollout error substantially (e.g., $+15.2$ px at $h{=}5$; Table~\ref{tab:cf1}), indicating that Driver-WM strictly relies on traffic cues for reactive forecasting.

\noindent\textbf{Dynamics core and training choices.}
Replacing the transition kernel with an RSSM/GRU-style recurrent dynamics degrades geometric fidelity and lowers semantic consistency, suggesting that self-attention is better suited for long-horizon conditional rollouts under high-dimensional context features. Skeleton pre-training yields only a limited effect relative to our main model and does not consistently improve semantics, indicating that the main gains of Driver-WM are not driven by warm-starting the skeleton predictor.

\noindent\textbf{Regularization and interface diagnostics.}
Introducing a KL bottleneck results in marginal changes across metrics, implying that the dominant performance driver in this setting is the causal injection structure rather than probabilistic regularization. Removing prompted external features consistently degrades both geometric and semantic metrics, highlighting the importance of the external feature interface for aligning driver dynamics with traffic context. Finally, removing the pose forecasting head retains competitive semantic performance on TCR but removes kinematic rollouts altogether; this diagnostic underscores the distinction between discriminative semantics and physically grounded forecasting, motivating kinematic prediction as the core foundation in Driver-WM.

\subsection{Interventions and Qualitative Analysis}
\label{subsec:cf_qual}

\begin{table*}[t]
\centering
\caption{\textbf{Controlled interventions on Driver-WM (main, test set).} We report absolute deviations ($\Delta$) in pixel space between the intervened rollouts and the factual (unintervened) rollouts. Larger deviations indicate higher sensitivity to the corresponding intervention. $h{=}5$ denotes the last predicted step.}
\label{tab:cf1}
\small
\setlength{\tabcolsep}{4.5pt}
\resizebox{\textwidth}{!}{%
\begin{tabular}{l l c c c c c}
\toprule
\textbf{Intervention $do(\cdot)$} &
\textbf{Target} &
$\Delta$All &
$\Delta$HM &
$\Delta h{=}5$ &
$\Delta$Head &
$\Delta$Hands \\
\midrule
\rowcolor{gray!10}
Factual (Driver-WM, main) & None & 0.000 & 0.000 & 0.000 & 0.000 & 0.000 \\
\midrule
$do(\text{Ext}=\text{Swap\_clip})$ & Cross-video content swap & 5.363 & 4.794 & 8.042 & 5.462 & 3.502 \\
$do(\text{Ext}=\emptyset)$ & Remove out-cabin features & 12.953 & 8.691 & 15.208 & 7.385 & 12.937 \\
$do(\text{Ext}=\text{Shift\_large})$ & Large temporal offset & 0.793 & 0.892 & 1.009 & 0.661 & 0.018 \\
$do(\text{Ext}=\text{DropView})$ & Drop one external view & 0.419 & 0.100 & 0.478 & 0.392 & 0.265 \\
\midrule
$do(\lambda_{\mathrm{CA}}{=}0)$ / $do(g{=}0)$ & Disable pathway & 89.641 & 62.232 & 116.751 & 36.783 & 95.666 \\
$do(\lambda_{\mathrm{CA}}{=}1)$ / $do(g{=}1)$ & Force injection & 26.733 & 11.480 & 42.170 & 15.861 & 26.727 \\
$do(\lambda_{\mathrm{CA}}{=}2)$ & Over-injection & 43.015 & 23.959 & 52.142 & 33.909 & 43.695 \\
\bottomrule
\end{tabular}%
}
\end{table*}

\begin{figure*}[t]
\centering
\includegraphics[width=\textwidth]{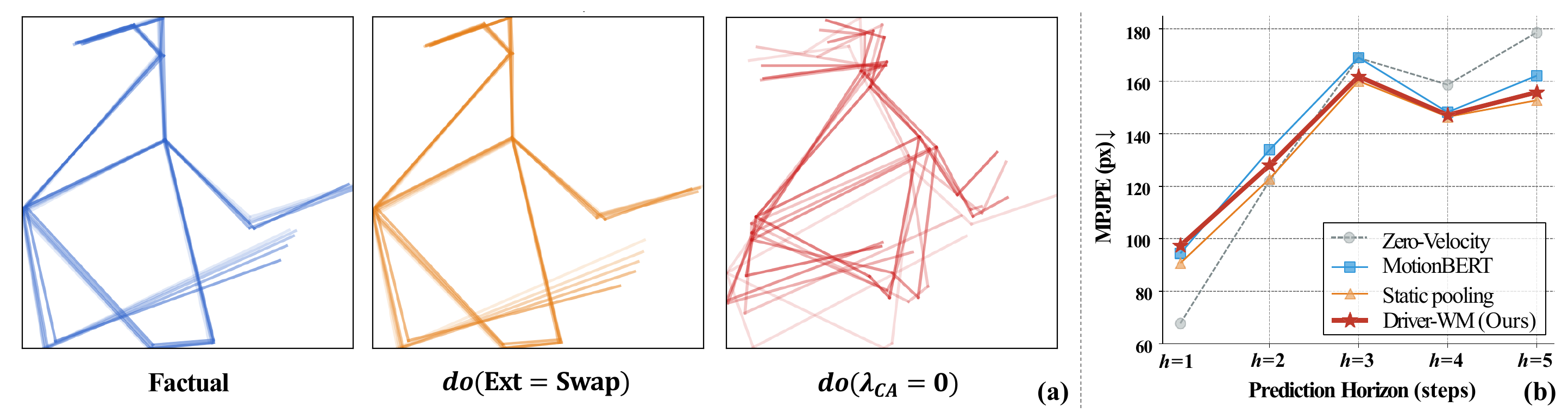}
\caption{\textbf{Mechanism and dynamics.}
(a) \textbf{Controlled interventions:} On the same clip, swapping the out-cabin context or disabling injection ($\lambda_{\mathrm{CA}}{=}0$) alters reactive hand motion; frames are aligned to the maximal injection step.
(b) \textbf{High-Motion tail:} Horizon-wise MPJPE shows the zero-velocity baseline degrades with horizon, while Driver-WM substantially reduces long-horizon error compared to motion-only baselines.}
\label{fig:cf_qual}
\end{figure*}

\begin{figure*}[t]
\centering
\includegraphics[width=\textwidth]{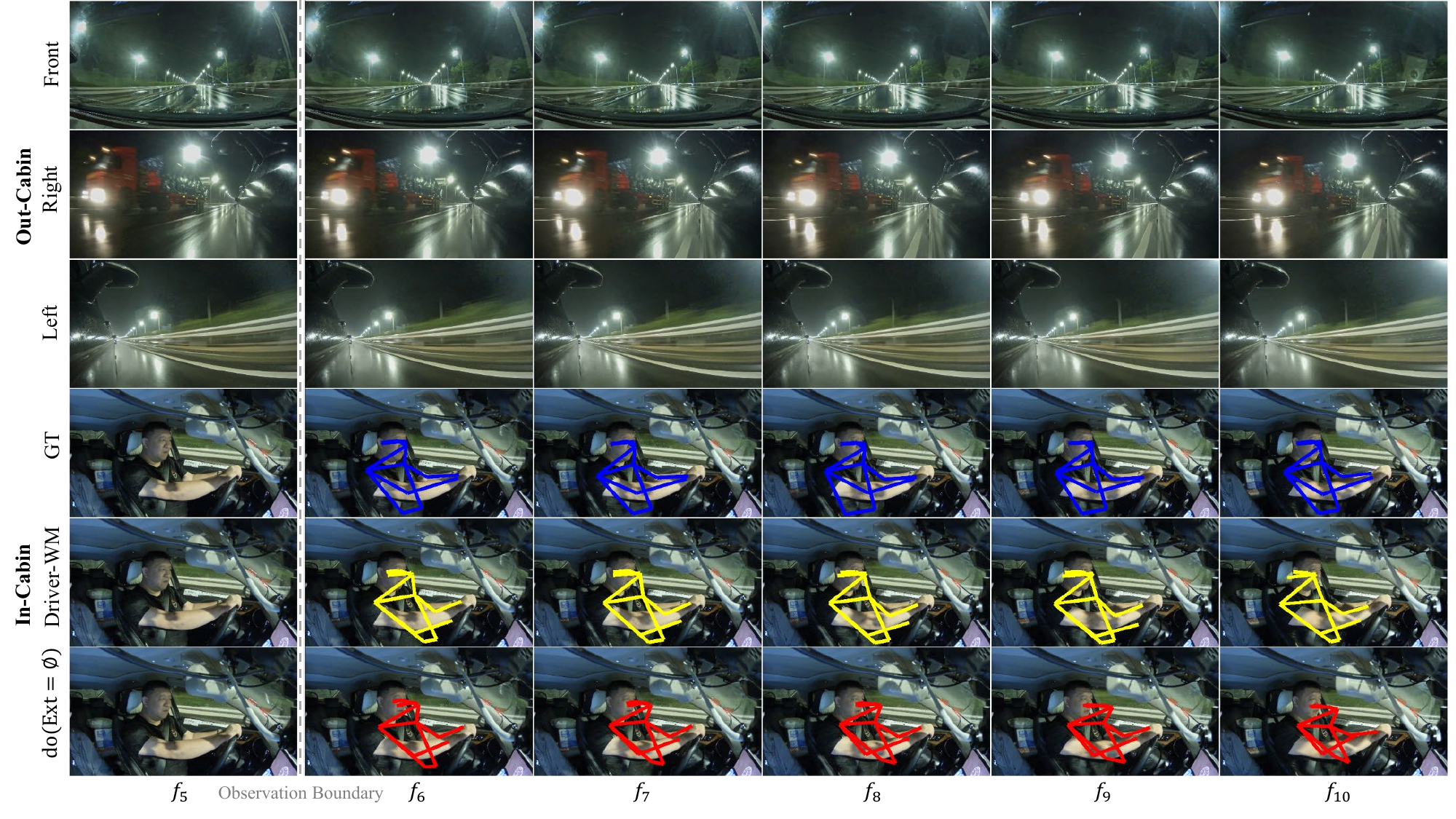}
\caption{\textbf{Qualitative results of driver dynamics rollout and causal interventions.} Setting: $5{\rightarrow}5$ rollout ($T_{\mathrm{obs}}{=}5$ frames observed, $T_{\mathrm{pred}}{=}5$ frames predicted). We compare the factual rollout of Driver-WM against the intervened rollout without external traffic context ($do(\mathrm{Ext}{=}\emptyset)$).
To avoid any video-generation artifacts, we directly overlay the 5-step predicted skeletons ($f_6{\sim}f_{10}$) onto the corresponding ground-truth in-cabin frames in pixel space.
The intervention yields a clearly different hand/upper-body kinematic trajectory in the future window.}
\label{fig:filmstrip}
\end{figure*}

We perform post-hoc test-time interventions on the trained Driver-WM (main) to assess (i) whether external context is effectively used and (ii) whether the injection pathway is necessary and controllable. Specifically, we intervene on the external stream (\texttt{Ext}) via cross-video content swapping, large temporal shifting, view dropping, or complete removal; or we intervene on the injection pathway by either clamping the learned gate $g$ to a constant or bypassing it with a scalar override $\lambda_{\mathrm{CA}}$; $do(g{=}0/1)$ is equivalent to $do(\lambda_{\mathrm{CA}}{=}0/1)$ under clamping or bypass, while $\lambda_{\mathrm{CA}}>1$ serves as an over-injection stress test. We use interventions as mechanism probes, without making causal effect estimation claims.

\noindent\textbf{Pathway necessity and injection strength.}
Table~\ref{tab:cf1} shows that disabling the injection pathway ($\lambda_{\mathrm{CA}}{=}0$ or $g{=}0$) yields the largest factual-rollout deviation, especially at $h{=}5$ and on hand joints. Forcing unmodulated injection ($\lambda_{\mathrm{CA}}{\in}\{1,2\}$) also causes substantial deviations, indicating that external signals should be injected with a learned, time-varying intensity rather than applied unconditionally. Direct-GT errors in the Supplementary Material confirm the same ordering.

\noindent\textbf{Effects of external context.}
Both swapping external context (Swap\_clip) and removing it entirely (Ext=$\emptyset$) induce consistent deviations, with a larger degradation when the external stream is removed. The effect is more pronounced at longer horizons and on reactive joints (hands). In contrast, large temporal shifts and dropping a single view produce comparatively minor deviations in this setting, suggesting empirical robustness to coarse temporal misalignment and partial viewpoint loss. This is consistent with early mean-pooling of multi-view external latents before entering the dynamics core.

\noindent\textbf{Visualizing causal responses.}
Figures~\ref{fig:cf_qual} and \ref{fig:filmstrip} provide qualitative counterparts of Table~\ref{tab:cf1}. In particular, Fig.~\ref{fig:cf_qual} visualizes the degradation under $do(\lambda_{\mathrm{CA}}{=}0)$.
Fig.~\ref{fig:filmstrip} overlays the predicted factual and intervened skeleton trajectories onto the corresponding ground-truth in-cabin frames (future window $f_6{\sim}f_{10}$) to highlight the kinematic divergence induced by intervention.

% ============================================================
% 5. Conclusion
% ============================================================
\section{Conclusion}
\label{sec:conclusion} 

We present Driver-WM, a dual-stream autoregressive world model that forecasts in-cabin driver dynamics explicitly conditioned on out-cabin traffic context. The architecture unifies the prediction of geometric skeleton trajectories and auxiliary semantic factors via a time-causal, gated cross-attention pathway, which dynamically injects external environmental cues into the internal latent rollout. Evaluations on the AIDE benchmark expose a critical inertia effect in standard $\ell_2$ metrics, where motion-only baselines exploit static extrapolation to attain deceptively low errors but degrade substantially on reactive maneuvers. Driver-WM circumvents this mean-regression trap, achieving robust kinematic fidelity alongside strong semantic alignment with the driving scene. Controlled test-time interventions corroborate the structural design: disabling the causal injection causes severe rollout degradation, while unmodulated injection leads to destabilization. This confirms the necessity of learned, time-varying contextual gating. Overall, Driver-WM shows that driver forecasting is best treated as a traffic-grounded rollout problem, bridging recognition-oriented cabin monitoring and human-aware risk reasoning.

% ==========================================================
% TODO FINAL: 提交给 AUMOVIO 审核版本 / 最终 Camera-ready 录用版本 时使用
% ⚠️ 警告：3月5日提交给 ECCV 系统的盲审版本中，下边这段必须全部注释掉（隐藏）！
\section*{Acknowledgements}
This work is supported by A*STAR under the RIE2025 Industry Alignment Fund – Industry Collaboration Projects (IAF-ICP) Funding Initiative (Award: I2501E0045), as well as cash and in-kind contribution from the industry partner(s).
% ==========================================================

% ---- Bibliography ----
%
% BibTeX users should specify bibliography style 'splncs04'.
% References will then be sorted and formatted in the correct style.
%
\bibliographystyle{splncs04}
\bibliography{main}

%下边的附录是我自己加上去的，到时候放在补充材料看，这些提交的时候要全都隐藏，对之后arxiv是有用的
% --- 新开一页开始补充材料 ---
% 投稿和交conti审核的时候需要去掉
% ------------
\clearpage
\appendix  % 切换章节编号格式为 A, B, C...
% % 建议使用 \input 而不是 \include，因为 \input 更灵活且不会强制分页
% 这个是我自己建立的
\newpage
\appendix
\section*{\Large{
\textit{Appendix}}}
\vskip8pt
% \startcontents
% {
% \hypersetup{linkcolor=black}
% \printcontents{}{1}{}
% }
% \clearpage

\section{Discussions}
\label{supp:discussion}
\vspace{-5pt}

Towards a better understanding of this work, we supplement intuitive questions that may arise.
Note that the following list does \textit{not} indicate the manuscript was submitted to a previous venue or not.

\noindent\textbf{Q1:}\textbf{ \textit{Why does Driver-WM simulate the driver instead of  detecting attention or monitoring takeover readiness?}}

\smallskip
A driver monitor system can tell us whether the driver is currently attentive, but it cannot tell us how the driver will respond to our next maneuver. In L2/L3 driving, the vehicle’s actions directly influence driver behavior—smooth deceleration, sudden braking, or lane changes all change how quickly and safely a driver re-engages. Instead, Driver-WM explicitly simulates this interaction. 
Predicting structured driver embodiment conditioned on external views provides driver-response cues for downstream risk analysis, rather than only reacting after a failure is detected.

\medskip
\noindent\textbf{Q2:} \textbf{\textit{Why can’t we formulate Driver-WM predicting takeover time or a readiness score?}}

\smallskip
Predicting driver monitoring signals provides only summarized supervision. Two drivers with the same predicted takeover time can behave very differently—one may smoothly regain control, while another may hesitate or overcorrect. Faced with the hidden behavior multi-modalities underlying future monitoring signals, our Driver-WM predicts the structure of the response: posture, hand movement, attention semantics, and their timing. These structured rollouts can be used by future human-aware planning modules to compare candidate maneuvers through predicted driver-response cues, rather than relying only on a scalar disengagement threshold.

\medskip
\noindent\textbf{Q3:} \textbf{\textit{What capabilities does Driver-WM unlock that are currently unavailable in L2/L3?}}

\smallskip
L2 and L3 systems often rely on conservative rules: warn early, disengage quickly, or brake when uncertainty rises. Such rules are safe but may reduce comfort and usable autonomy. Driver-WM does not output planning actions; instead, it provides short-horizon driver-response rollouts that can inform future risk modules. These cues may support conservative maneuver selection, takeover-timing analysis, or human-machine interface adaptation when combined with calibrated vehicle states and downstream validation. As autonomy capabilities expand, a structured driver-response model can offer a common interface across L2/L3 settings without redesigning the in-cabin perception stack.

\section{Additional Implementation Details}
\label{sec:appendix_details}

This appendix provides additional definitions and implementation details omitted from the main paper due to space constraints.%, including (i) the causal context operator and ablation variants, (ii) multi-view out-cabin fusion with view embeddings, (iii) skeleton representation and teacher distillation protocol, and (iv) the explicit form of the cabin/seat feasibility prior and the supervised probabilistic bottleneck.

\subsection{AIDE Dataset and High-Motion Subsets}
\label{app:aide_dataset}

While the fundamental task protocol and temporal sampling strategy ($5 \rightarrow 5$ causal rollout at $\sim$3.3\,Hz) are detailed in the main text, we provide additional specific dataset configurations necessary for strict reproducibility.

\paragraph{Data Partition and Keypoint Topology.}
We strictly adhere to the official AIDE evaluation split without performing subject-held-out re-splitting. This yields an exact distribution of 1,884 clips for training, 405 for validation, and 609 for testing. For kinematic modeling, we utilize the official HALPE-136 2D skeleton annotations. This topology is densely structured with $K{=}136$ joints, specifically comprising 26 body joints, 68 facial landmarks, and 42 hand joints ($21$ per hand). Explicitly modeling the 42 hand joints is critical for analyzing driver reactions to out-cabin traffic events. The specific semantic class definitions used for the auxiliary regularizers are listed in Table~\ref{tab:aide_labelsets}.

\paragraph{High-Motion (HM) Subset Definition.}
To rigorously evaluate the model's capability in safety-critical, reactive scenarios, we define a High-Motion (HM) tail on the test set. This subset is established by ranking clips according to a ground-truth motion score, computed strictly on the future window $\{t=T_{\mathrm{obs}}+1,\dots,T\}$. Let $\mathbf{s}^{gt}_{t}\in\mathbb{R}^{K\times 2}$ denote the ground-truth keypoints in original pixel coordinates. The motion score is defined as the mean per-joint frame-to-frame displacement:
\begin{equation}
\label{eq:hm_motion_score}
\mathrm{motion\_score}
=
\frac{1}{(T_{\mathrm{pred}}-1)K}
\sum_{t=T_{\mathrm{obs}}+1}^{T-1}
\sum_{k=1}^{K}
\left\|
\mathbf{s}^{gt}_{t+1,k}-\mathbf{s}^{gt}_{t,k}
\right\|_2.
\end{equation}
We rank all 609 test clips by Eq.~\eqref{eq:hm_motion_score} and select the top 10\%. This yields a fixed subset of $N_{\mathrm{HM}}{=}60$ clips, corresponding to a motion score threshold of $\ge 72.05$ pixels. The HM subset is universally shared across all horizons and compared baselines to ensure a completely fair evaluation of reactive dynamics.

\begin{table}[h]
\centering
\caption{AIDE task label sets (class order follows the official AIDE benchmark visualization).}
\label{tab:aide_labelsets}
\small
\setlength{\tabcolsep}{4pt}
\renewcommand{\arraystretch}{1.05}
\resizebox{\columnwidth}{!}{%
\begin{tabular}{ll}
\toprule
\textbf{Task} & \textbf{Classes} \\
\midrule
DBR & Normal Driving; Looking Around; Making Phone; Body Movement; Talking; Smoking; Dozing Off \\
DER & Peace; Anxiety; Weariness; Happiness; Anger \\
TCR & Smooth Traffic; Waiting; Traffic Jam \\
VCR & Forward Moving; Parking; Turning; Backward Moving; Changing Lane \\
\bottomrule
\end{tabular}%
}
\end{table}

\subsection{Prompt-Guided Feature Extraction with Frozen VLM}
\label{app:prompt_feature_extraction}

To construct the dual-stream external and internal context for Driver-WM, we pre-extract frame-level features using a frozen vision-language model (Qwen-VL series, 2B). The extraction is performed offline, operating strictly without access to any ground-truth dataset annotations or multi-task labels to prevent information leakage.

\paragraph{Fixed Text Prompt.}
To explicitly steer the frozen encoder toward interaction-aware semantics rather than generic object recognition, we pair every input frame with a fixed natural-language prompt. The exact prompt string used across all splits and camera views is:
\begin{quote}
``\textit{Extract multimodal features that capture: (A) external traffic cues (vehicles, signals, road agents), (B) how these cues act as triggers for driver motion responses, and (C) the short-term evolution (1-2 seconds) of driver behavior in response to the changing road context.}''
\end{quote}
This single, static instruction acts as a textual prior, forcing the model to encode causal and dynamic elements within the scene into its hidden states.

\paragraph{Encoding and Pooling Strategy.}
For each synchronized timestamp, the RGB frame and the fixed prompt are processed through the VLM. We extract the hidden states from the final layer and apply mean pooling over the output sequence dimension. This yields a single $D=2048$ dimensional feature vector per frame. For each 3-second clip, the extracted frame-level features are stacked into a $T \times D$ tensor (where $T{=}10$) and cached offline for each respective view.

\paragraph{Usage in Driver-WM.}
The cached feature tensors ($\mathbf{f}^{\text{in}}_{t}$ and $\mathbf{f}^{\text{out}}_{t,v}$) are treated as fixed observations strictly for the historical window ($t \le T_{\mathrm{obs}}$). During the training and evaluation of Driver-WM, gradients are not propagated into the VLM. For the future prediction window ($t > T_{\mathrm{obs}}$), the external context is advanced autoregressively in the latent space. The internal rollout transition at time step $t$ accesses only the causally valid external history up to $t$ ($\hat{\bar{\mathbf{Z}}}^{\text{ext}}_{\le t}$), rigorously preserving zero-lookahead temporal causality.

\subsection{Pretraining Strategies and Warm-start Variants}
\label{app:pretraining_details}

To disentangle spatial geometric decoding from temporal causal dynamics, we decouple the pretraining of the Driver-WM components into two isolated streams, corresponding to the warm-start variants evaluated in our ablation studies.

\paragraph{Skeleton Decoder Warm-start (S-Pre).}
Rather than using teacher-forced visual latents, we warm-start the spatial decoding head ($D_{\text{skel}}$) via a Coordinate Auto-Encoder formulation. Specifically, the ground-truth 2D skeleton sequences $\mathbf{s}_{gt}$ are projected into a proxy latent space $\mathbf{z}_{\text{skel}}$ using a temporary lightweight encoder, and subsequently reconstructed by the ST-GCN decoder. This phase is optimized solely using the geometric regression loss $\mathcal{L}_{\text{skel}}$ (without causal injection or auxiliary semantics). 
As analyzed in the main text, the resulting downstream gains on the AIDE benchmark are modest. This diagnostic confirms that the dominant error in long-horizon High-Motion rollouts stems from context-conditioned internal transitions rather than spatial decoding capacity; once the temporal transition core is robustly trained, warm-starting the spatial decoder provides limited additional benefits.

\paragraph{Internal Dynamics Pretraining on Auxiliary Data.}
We optionally warm-start the internal transition core $f^{\text{int}}_{\theta}$ on Drive\&Act using latent multi-step prediction without external context. Given an internal latent window $\mathbf{Z}^{\text{int}}_{1:L}$, we take the last state $\mathbf{z}^{\text{int}}_{L}$ and autoregressively predict $K$ future steps $\{\hat{\mathbf{z}}^{\text{int}}_{L+k}\}_{k=1}^{K}$. The objective is configured as an $L_2$ loss, cosine loss, or a mixed direction--magnitude loss, and feature normalization is controlled by the pretraining configuration. The resulting weights are used to initialize $f^{\text{int}}_{\theta}$ before fine-tuning on AIDE with the full Driver-WM model.

\subsection{Strict Zero-Lookahead Verification}
\label{supp:no_leak}

To rule out any potential future information path during evaluation, we conduct numerical invariance tests on the predicted skeletal tensors. 
Specifically, at each rollout horizon $h\in\{1,\dots,5\}$, we perturb the future portion beyond the causal cutoff either by (i) zeroing the suffix (prefix-equivalence proxy), or (ii) replacing it with randomized noise (suffix randomization).
Table~\ref{tab:no_leak} reports the maximum absolute difference (max abs diff) between the perturbed and the original predictions. 
All differences remain below $3\times10^{-7}$ across horizons, numerically confirming strict time-causality.

\begin{table}[t]
\centering
\caption{\textbf{Numerical invariance tests for strict time-causality (no leakage).} 
We report max abs diff on predicted skeleton tensors under two perturbations applied to the unobservable future external suffix.}
\label{tab:no_leak}
\setlength{\tabcolsep}{6pt}
\resizebox{\textwidth}{!}{%
\begin{tabular}{lccccc}
\toprule
\textbf{Test (max abs diff)} & \textbf{h1} & \textbf{h2} & \textbf{h3} & \textbf{h4} & \textbf{h5} \\
\midrule
Prefix-equivalence (suffix-zero) & $2.980232{\times}10^{-7}$ & $2.980232{\times}10^{-7}$ & $2.980232{\times}10^{-7}$ & $2.980232{\times}10^{-7}$ & $2.980232{\times}10^{-7}$ \\
Suffix randomization (strict future) & $2.980232{\times}10^{-7}$ & $2.980232{\times}10^{-7}$ & $2.980232{\times}10^{-7}$ & $2.980232{\times}10^{-7}$ & $2.980232{\times}10^{-7}$ \\
\bottomrule
\end{tabular}
}
\end{table}

We also rerun the same forward pass under fixed seed and evaluation mode, obtaining a self-consistency max abs diff of $3.576279{\times}10^{-7}$.

\subsection{Rationale for the ST-GCN Decoder.}
Recent motion forecasters (e.g., MotionBERT, SiMLPe) demonstrate strong performance as coordinate-to-coordinate sequence models. However, we deliberately adopt a lightweight ST-GCN as our skeleton decoder. This choice aligns with the latent world model paradigm: the complex temporal dynamics and context-conditioned reactions are fully resolved within the latent transition core. Consequently, the decoding head's primary role is spatial rather than temporal---projecting the rolled-out semantic latent state back to physically plausible joint coordinates. ST-GCN provides the explicit skeletal topology (kinematic graph) necessary for this spatial decoding. Employing a heavy spatio-temporal forecaster as the head would introduce redundant temporal modeling and confound the evaluation, blurring whether performance gains stem from our Gated Causal Injection or simply a high-capacity pose module.

% ------------------------------------------------------------------
\subsection{Causal Context Operator and Coupling Variants}
\label{app:ctx_gate}

\paragraph{Implementation of $\mathrm{Ctx}_\theta$.}
We implement $\mathrm{Ctx}_\theta$ as a lightweight Perceiver-style cross-attention module. Temporal causality is enforced by construction: at rollout step $t$, the module only receives truncated prefixes $\hat{\mathbf{Z}}^{\text{int}}_{\le t}$ and $\bar{\mathbf{Z}}^{\text{ext}}_{\le t}$, without any explicit attention masking.

Concretely, we first mean-pool the internal history and map it through an MLP to produce a small set of latent queries. These queries attend to keys and values obtained from the external prefix (with low-rank bottleneck projections). The attention output is projected to a global context vector and broadcast-added to the internal sequence, yielding a context-modulated history. We take the last context-modulated state as the summary $\mathbf{m}_t\in\mathbb{R}^{D}$ used in 
\begin{equation}
\label{eq:gated_fusion_supp}
\mathbf{g}_t=\sigma\!\left(\mathrm{MLP}_{g}\!\left(\hat{\bar{\mathbf{z}}}^{\text{ext}}_{t}\right)\right),
\qquad
\hat{\mathbf{z}}^{\text{int}}_{t+1} 
= (\mathbf{1}-\mathbf{g}_t)\odot \tilde{\mathbf{z}}^{\text{int}}_{t+1}
\;+\;
\mathbf{g}_t\odot \mathbf{m}_{t}.
\end{equation}

\paragraph{Gate input variants.}
The default gate uses only the current pooled external latent:
\begin{equation}
\mathbf{g}_t=\sigma\!\left(\mathrm{MLP}_{g}\!\left(\bar{\mathbf{z}}^{\text{ext}}_{t}\right)\right)\in(0,1)^D.
\end{equation}
When explicit external semantic attributes are available, we concatenate their embedding:
\begin{equation}
\mathbf{g}_t=\sigma\!\left(\mathrm{MLP}_{g}\!\left([\bar{\mathbf{z}}^{\text{ext}}_{t},\,\mathbf{e}_{c,t}]\right)\right).
\end{equation}

% 因为实际上从来就没有用到residual gate的可以展示的结果，所以这部分可以先隐藏。
% \paragraph{Intervention setup.}
% All reported $\lambda_{\mathrm{CA}}$ interventions assume the optional residual gate is disabled to isolate the primary contextual pathway.

\paragraph{Definition of the Maximal Injection Step.}
For the qualitative visualization of causal responses in the main text, the ``maximal injection step'' is defined as the time step $t$ that yields the maximum mean activation across all dimensions of the vector gate $\mathbf{g}_t$. Specifically, $t_{\mathrm{max}} = \arg\max_{t} (\frac{1}{D}\sum_{d=1}^{D} g_{t, d})$. This provides a scalar proxy to temporally align frames where the external context exerts the strongest overall influence on the internal dynamics.

% \paragraph{Ablation: instantaneous interaction.}
% To remove long-range dependency modeling, we replace causal cross-attention with a local interaction:
% \begin{equation}
% \mathbf{m}_t = \mathrm{MLP}_{\text{inst}}\!\left([\bar{\mathbf{z}}^{\text{ext}}_{t},\,\hat{\mathbf{z}}^{\text{int}}_{t}]\right).
% \end{equation}

% \paragraph{Ablation: perturbation-based coupling.}
% We also evaluate a perturbation coupling that injects a gated external-dependent residual into the internal transition:
% \begin{equation}
% \Delta \mathbf{p}_t=\mathrm{MLP}_{p}([\bar{\mathbf{z}}^{\text{ext}}_{t},\,\hat{\mathbf{z}}^{\text{int}}_{t}])\in\mathbb{R}^{D},
% \qquad
% \hat{\mathbf{z}}^{\text{int}}_{t+1} = \tilde{\mathbf{z}}^{\text{int}}_{t+1} + \mathbf{g}_t\odot \Delta \mathbf{p}_t.
% \end{equation}

% ------------------------------------------------------------------
\subsection{Multi-view Out-cabin Fusion with View Embeddings}
\label{app:multiview}

AIDE provides three out-cabin cameras (front/left/right) and one in-cabin camera.
Let $\mathbf{f}^{\text{out},v}_t=E_{\text{vlm}}(\mathbf{o}^{\text{out},v}_t)\in\mathbb{R}^{D}$ denote the Qwen3-VL feature from out-cabin view $v\in\{\text{front},\text{left},\text{right}\}$.
We first inject a learnable view embedding, then fuse the out-cabin views into a single external latent.
Our default fusion aligns with the main text by first extracting the view-conditioned representations and subsequently performing mean pooling:
\begin{equation}
\bar{\mathbf{f}}^{\text{out},v}_t = \mathbf{f}^{\text{out},v}_t + \mathbf{e}_{\text{view}}(v),
\qquad
\mathbf{z}^{\text{ext}}_{t,v} = \bar{\mathbf{f}}^{\text{out},v}_t,
\qquad
\bar{\mathbf{z}}^{\text{ext}}_t = \frac{1}{3}\sum_{v}\mathbf{z}^{\text{ext}}_{t,v}.
\end{equation}
For the in-cabin stream (single view), we use:
\begin{equation}
\mathbf{z}^{\text{int}}_t = \mathbf{f}^{\text{in}}_t + \mathbf{e}_{\text{view}}(v_{\text{in}}).
\end{equation}
(Alternative fusion schemes, e.g., learned attention pooling over views, can be plugged in, but are not required for the main results.)

% ------------------------------------------------------------------
\subsection{Skeleton Representation and Teacher-based Distillation}
\label{app:skeleton_teacher}

\paragraph{Skeleton topology and kinematic tree.}
We represent the driver using a HALPE-136 style skeleton with $K{=}136$ keypoints.
Let $\mathcal{E}$ denote the set of kinematic edges (bone pairs) used for bone-length regularization (Eq.~\eqref{eq:bone_loss_supp}).
In our implementation, $\mathcal{E}$ follows the standard HALPE whole-body tree (body+hands), and we keep the joint ordering consistent across all splits.

\paragraph{Coordinate normalization and missing-joint masking.}
We compute skeleton losses in normalized image coordinates.
Given a keypoint $(x,y)$ in pixel coordinates for a frame of size $(W,H)$, we normalize it as:
\begin{equation}
\tilde{x}=x/W,\qquad \tilde{y}=y/H,
\end{equation}
and compute MPJPE/PCK on the normalized coordinates.
To handle occlusions and low-confidence detections from the teacher, we use a per-joint confidence mask $\mathbf{m}_{t}\in\{0,1\}^{K}$:
\begin{equation}
\mathcal{L}_{\text{skel}}=
\sum_{t}\frac{1}{\sum_{i}m_{t,i}+\epsilon}\sum_{i=1}^{K} m_{t,i}\left\|\hat{\mathbf{s}}_{t,i}-\mathbf{s}_{t,i}\right\|_2^2,
\end{equation}
where $m_{t,i}=1$ if the teacher confidence exceeds a threshold and $0$ otherwise.

% ------------------------------------------------------------------
\subsection{Physical and Kinematic Priors}
\label{app:phys_priors}

To ensure that the autoregressive rollouts remain anatomically and geometrically plausible over long horizons, we supplement the primary regression loss with a composite physical prior:
\begin{equation}
\label{eq:total_phys_loss_supp}
\mathcal{L}_{\text{phys}} = \lambda_{\text{bone}}\mathcal{L}_{\text{bone}} + \lambda_{\text{smooth}}\mathcal{L}_{\text{smooth}} + \lambda_{\text{seat}}\mathcal{L}_{\text{seat}}.
\end{equation}

\paragraph{Kinematic Tree Consistency.}
For variants utilizing 3D keypoint lifting or supervision, we enforce constant bone lengths across the kinematic tree $\mathcal{E}$ (e.g., the HALPE-136 whole-body topology). The bone-length consistency loss is defined as:
\begin{equation}
\label{eq:bone_loss_supp}
\mathcal{L}_{\text{bone}} = \sum_{t=1}^{T} \sum_{(i,j) \in \mathcal{E}} \left| \| \hat{\mathbf{s}}_{t,i} - \hat{\mathbf{s}}_{t,j} \|_2 - \| \mathbf{s}_{t,i} - \mathbf{s}_{t,j} \|_2 \right|,
\end{equation}
where $\mathbf{s}_{i,j}$ denote the ground-truth joint coordinates.

\paragraph{Temporal Smoothness.}
To mitigate high-frequency jitter in self-conditioned rollouts, we penalize first- and second-order temporal differences:
\begin{equation}
\label{eq:smooth_loss_supp}
\mathcal{L}_{\text{smooth}} = \sum_{t} \| \Delta \hat{\mathbf{s}}_{t} - \Delta \mathbf{s}_{t} \|_2^2 + \sum_{t} \| \Delta^2 \hat{\mathbf{s}}_{t} \|_2^2,
\end{equation}
where $\Delta \hat{\mathbf{s}}_{t} = \hat{\mathbf{s}}_{t+1} - \hat{\mathbf{s}}_{t}$ and $\Delta^2 \hat{\mathbf{s}}_{t} = \hat{\mathbf{s}}_{t+2} - 2\hat{\mathbf{s}}_{t+1} + \hat{\mathbf{s}}_{t}$.

\paragraph{Cabin ROI Feasibility.}
We define a soft Region-of-Interest (ROI) penalty to discourage key joints $\mathcal{J}$ (e.g., head, hips) from drifting outside anatomically plausible cabin boundaries. Given $\text{ROI} = [x_{\min}, x_{\max}] \times [y_{\min}, y_{\max}]$, the loss is formulated as:
\begin{equation}
\begin{split}
\mathcal{L}_{\text{seat}} = \sum_{t} \sum_{j \in \mathcal{J}} \Big( & \text{ReLU}(x_{t,j} - x_{\max}) + \text{ReLU}(x_{\min} - x_{t,j}) \\
& + \text{ReLU}(y_{t,j} - y_{\max}) + \text{ReLU}(y_{\min} - y_{t,j}) \Big).
\end{split}
\end{equation}

% ------------------------------------------------------------------
\subsection{Stochastic Transition Core and KL Bottleneck}
\label{app:kl_closedform}

In our probabilistic ablation variants, the transition operator $\mathcal{F}_\theta$ is factorized into a stochastic internal predictor $f_\theta$ followed by conditioned contextual injection.

\paragraph{Reparameterized Transition.}
The predictor $f_\theta$ outputs the parameters of a diagonal Gaussian distribution representing the candidate next state:
\begin{equation}
\label{eq:int_transition_supp}
(\boldsymbol{\mu}_{t+1}, \log \boldsymbol{\sigma}_{t+1}) = f_{\theta}(\hat{\mathbf{z}}^{\text{int}}_{t}), \quad \tilde{\mathbf{z}}^{\text{int}}_{t+1} = \boldsymbol{\mu}_{t+1} + \boldsymbol{\epsilon} \odot \boldsymbol{\sigma}_{t+1}, \quad \boldsymbol{\epsilon} \sim \mathcal{N}(\mathbf{0}, \mathbf{I}).
\end{equation}
For the deterministic Driver-WM used in our main results, we set $\boldsymbol{\sigma}_{t+1} = \mathbf{0}$.

\paragraph{Supervised KL Regularization.}
When stochastic modeling is enabled, we regularize the predicted density $q_{\theta}(\tilde{\mathbf{z}}^{\text{int}}_{t+1} \mid \hat{\mathbf{z}}^{\text{int}}_{t})$ with a \emph{conditional prior} $p(\mathbf{z}^{\text{int}}_{t+1}) = \mathcal{N}(\mathbf{z}^{\text{int}}_{t+1}, \mathbf{I})$ centered at the ground-truth latent. This supervised probabilistic bottleneck enforces latent consistency via:
\begin{equation}
\label{eq:kl_cond}
\mathcal{L}_{\text{KL}} = \sum_{t=1}^{T-1} \text{KL}\left( \mathcal{N}(\boldsymbol{\mu}_{t+1}, \text{diag}(\boldsymbol{\sigma}^2_{t+1})) \| \mathcal{N}(\mathbf{z}^{\text{int}}_{t+1}, \mathbf{I}) \right).
\end{equation}
The KL term admits the following closed-form solution:
\begin{equation}
\mathcal{L}_{\text{KL}} = \frac{1}{2} \sum_{t=1}^{T-1} \sum_{i=1}^{D} \left( \sigma_{t+1,i}^2 + (\mu_{t+1,i} - z^{\text{int}}_{t+1,i})^2 - 1 - \log \sigma_{t+1,i}^2 \right).
\end{equation}

\paragraph{Unified Optimization Objective.}
The total training objective integrates the geometric rollout loss, latent consistency, and the aforementioned priors:
\begin{equation}
\label{eq:total_loss_supp}
\mathcal{L} = \lambda_{\text{lat}}\mathcal{L}_{\text{lat}} + \lambda_{\text{skel}}\mathcal{L}_{\text{skel}} + \lambda_{\text{aux}}\mathcal{L}_{\text{aux}} + \beta \mathcal{L}_{\text{KL}} + \lambda_{\text{phys}}\mathcal{L}_{\text{phys}}.
\end{equation}
In practice, model selection is driven by $h{=}1\dots5$ MPJPE, while $\mathcal{L}_{\text{KL}}$ and $\mathcal{L}_{\text{phys}}$ serve as controllable regularizers to stabilize long-term kinematic rollouts.
% ------------------------------------------------------------------
\subsection{Additional Training Details (Fill-in Table)}
\label{app:impl_table}

For completeness and reproducibility, we summarize key hyperparameters in Table~\ref{tab:impl_details}.

\begin{table}[t]
\centering
\small
\begin{tabular}{l l}
\toprule
\textbf{Hyperparameter} & \textbf{Setting} \\
\midrule
\multicolumn{2}{c}{\textit{Architecture Details}} \\
\midrule
VLM feature dim $D$ & 2048 (Qwen3-VL 2B, frozen) \\
Latent interface: identity (view embedding only) & TransitionMLP \\
Causal Temporal core & 1 layer, 4 heads, $D_{ff}{=}4096$ \\
Gated Cross-Attn & 1 layer, 4 heads \\
\midrule
\multicolumn{2}{c}{\textit{Data Processing \& Setup}} \\
\midrule
Sampling strategy & 10 frames uniformly sampled over 3.0s \\
$T_{\mathrm{obs}}$ / $T_{\mathrm{pred}}$ & ~1.5s (5 frames) / ~1.5s (5 frames) \\
Target pose format & HALPE-136 (136 keypoints) \\
\midrule
\multicolumn{2}{c}{\textit{Optimization Setup}} \\
\midrule
Optimizer & AdamW ($\beta_1=0.9, \beta_2=0.999$) \\
Learning rate schedule & Cosine annealing, base LR $5\times 10^{-5}$ \\
Weight decay & $1\times 10^{-5}$ \\
Batch size / Max epochs & 32 / 100 (Best epoch typically $\sim$78) \\
Loss weights & $\lambda_{\text{skel}}{=}1.0$, $\lambda_{\text{sem}}{=}1.0$ (equal weighting) \\
\bottomrule
\end{tabular}
\caption{Implementation details of the proposed Driver-WM.}
\label{tab:impl_details}
\end{table}

\clearpage

\section{Additional Results}
\label{app:add_results}

% ==========================================================
% 补充材料：全量实验结果总表 (Seed 42)
% ==========================================================
\begin{table*}[h]
\centering
\caption{\textbf{Comprehensive performance summary of all evaluated models (seed 42).} We report horizon-averaged geometric errors (MPJPE in px and diagonal-normalized \%), PCK, and semantic metrics (Accuracy and Macro-F1). This supplementary table provides exhaustive metrics across all classification and forecasting tasks to manifest the multi-task capacity. For the \textit{no pose head} variant, kinematic forecasting metrics are not applicable (marked as ``--'') as the skeletal decoder is removed. For motion-only \& offline baselines, externally grounded semantics (TCR/VCR) are inapplicable and marked as ``--''.}
\label{tab:supp_seed42_full}
\small 
\setlength{\tabcolsep}{3.2pt} 
\resizebox{\textwidth}{!}{%
\begin{tabular}{c l cc cc cc cc cc cc}
\toprule
\textbf{Rank} & \textbf{Model} & \multicolumn{2}{c}{\textbf{All MPJPE}} & \multicolumn{2}{c}{\textbf{HM MPJPE}} & \textbf{PCK} & \textbf{PCK} & \textbf{DBR} & \textbf{DER} & \textbf{TCR} & \textbf{VCR} \\
& & \textbf{(px)} & \textbf{(\%)} & \textbf{(px)} & \textbf{(\%)} & \textbf{@0.05} & \textbf{@0.10} & \textbf{(Acc/F1)} & \textbf{(Acc/F1)} & \textbf{(Acc/F1)} & \textbf{(Acc/F1)} \\
\midrule
1 & Zero-Velocity & 52.89 & 2.40 & 139.19 & 6.32 & 85.95\% & 93.61\% & 40.72 / 8.27 & 59.61 / 14.94 & -- / -- & -- / -- \\
2 & Static pooling (no rollout) & 68.50 & 3.11 & 134.56 & 6.11 & 72.55\% & 89.10\% & 69.62 / 54.04 & 73.56 / 60.07 & 65.52 / 26.39 & 53.86 / 14.00 \\
3 & KL bottleneck & 70.56 & 3.20 & 136.50 & 6.20 & 71.71\% & 88.33\% & 75.37 / 69.89 & 77.83 / 71.54 & 92.45 / 88.84 & 81.61 / 64.68 \\
4 & +Skeleton-Pre & 70.92 & 3.22 & 136.04 & 6.18 & 72.09\% & 88.30\% & 74.06 / 69.55 & 81.61 / 75.65 & 92.28 / 88.33 & 82.92 / 70.24 \\
5 & Driver-WM (main) & 71.47 & 3.24 & 138.03 & 6.27 & 71.66\% & 88.16\% & 73.23 / 68.07 & 79.47 / 72.61 & 93.10 / 90.15 & 82.59 / 68.34 \\
6 & w/o ext context & 71.74 & 3.26 & 136.94 & 6.21 & 71.13\% & 88.20\% & 73.40 / 71.33 & 74.22 / 63.56 & 65.52 / 26.39 & 53.86 / 14.00 \\
7 & Non-causal (bidir) & 72.15 & 3.27 & 136.50 & 6.20 & 71.39\% & 88.79\% & 73.40 / 73.35 & 81.44 / 74.46 & 92.12 / 88.65 & 81.94 / 68.82 \\
8 & Driver-WM (full pretrained) & 72.68 & 3.30 & 138.03 & 6.27 & 70.99\% & 88.25\% & 75.37 / 73.22 & 80.62 / 73.97 & 93.43 / 90.54 & 81.94 / 62.09 \\
9 & MotionBERT & 73.51 & 3.34 & 141.53 & 6.42 & 78.01\% & 92.94\% & 67.98 / 56.70 & 66.34 / 52.71 & -- / -- & -- / -- \\
10 & Offline Enc-Dec & 75.87 & 3.45 & 144.05 & 6.54 & 77.17\% & 93.07\% & 67.32 / 53.82 & 64.70 / 50.68 & -- / -- & -- / -- \\
11 & no-prompt feat. & 76.54 & 3.47 & 142.08 & 6.45 & 68.28\% & 86.45\% & 73.89 / 68.88 & 79.31 / 69.92 & 85.39 / 76.87 & 77.18 / 58.59 \\
12 & RSSM/GRU dyn. & 78.42 & 3.56 & 136.46 & 6.19 & 67.48\% & 86.69\% & 70.77 / 67.15 & 79.80 / 72.29 & 92.12 / 88.26 & 81.44 / 63.74 \\
13 & Cross-Attn only & 80.14 & 3.64 & 142.41 & 6.47 & 66.43\% & 85.75\% & 67.49 / 62.52 & 75.53 / 68.75 & 92.12 / 87.94 & 82.92 / 69.09 \\
14 & Late fusion & 82.59 & 3.75 & 143.31 & 6.51 & 65.07\% & 85.28\% & 67.98 / 45.87 & 68.80 / 49.10 & 92.78 / 87.85 & 82.92 / 65.97 \\
15 & Single-stream & 90.87 & 4.13 & 147.44 & 6.69 & 59.63\% & 82.05\% & 54.52 / 24.56 & 62.07 / 24.35 & 65.52 / 26.39 & 53.86 / 14.00 \\
16 & SiMLPe & 106.38 & 4.83 & 156.29 & 7.09 & 63.45\% & 86.29\% & 59.44 / 37.26 & 66.34 / 41.69 & -- / -- & -- / -- \\
17 & ST-GCN & 110.98 & 5.04 & 158.10 & 7.18 & 60.97\% & 84.88\% & 51.23 / 18.96 & 61.08 / 22.66 & -- / -- & -- / -- \\
--- & no pose head & -- & -- & -- & -- & -- & -- & 72.74 / 67.22 & 77.50 / 70.92 & 92.94 / 88.83 & 83.74 / 69.44 \\
\bottomrule
\end{tabular}%
}
\end{table*}

% ==============================================================================
% 补充材料：多Seed稳健性与方差分析表
% ==============================================================================

% \subsection{Multi-Seed Robustness Analysis}

\begin{table*}[h]
\centering
\caption{\textbf{Comprehensive multi-seed robustness analysis (3 seeds: 42, 123, 456).} We report the mean and standard deviation ($\mu \pm \sigma$) for all geometric and semantic metrics under the $5{\rightarrow}5$ rollout protocol. For motion-only baselines, externally grounded semantics (TCR/VCR) are physically inapplicable and marked as ``--''. $\dagger$: copy-last-frame baseline.}
\label{tab:supp_multi_seed}
\small
\setlength{\tabcolsep}{2.5pt}
\resizebox{\textwidth}{!}{%
\begin{tabular}{l cc cc cc cc cc cc cc}
\toprule
\textbf{Model} & 
\multicolumn{2}{c}{\textbf{All MPJPE}} & 
\multicolumn{2}{c}{\textbf{HM MPJPE}} & 
\multicolumn{2}{c}{\textbf{DBR}} & 
\multicolumn{2}{c}{\textbf{DER}} & 
\multicolumn{2}{c}{\textbf{TCR}} & 
\multicolumn{2}{c}{\textbf{VCR}} & 
\textbf{PCK} & \textbf{PCK} \\
& \textbf{(px)}$_{\downarrow}$ & \textbf{(\%)}$_{\downarrow}$ & \textbf{(px)}$_{\downarrow}$ & \textbf{(\%)}$_{\downarrow}$ & \textbf{Acc}$_{\uparrow}$ & \textbf{F1}$_{\uparrow}$ & \textbf{Acc}$_{\uparrow}$ & \textbf{F1}$_{\uparrow}$ & \textbf{Acc}$_{\uparrow}$ & \textbf{F1}$_{\uparrow}$ & \textbf{Acc}$_{\uparrow}$ & \textbf{F1}$_{\uparrow}$ & \textbf{@0.05}$_{\uparrow}$ & \textbf{@0.10}$_{\uparrow}$ \\
\midrule
\multicolumn{15}{l}{\textit{Motion-only \& offline baselines}} \\
Zero-Velocity$\dagger$ & 52.89$\pm$0.0 & 2.40$\pm$0.0 & 139.19$\pm$0.0 & 6.32$\pm$0.0 & 40.72$\pm$0.0 & 8.27$\pm$0.0 & 59.61$\pm$0.0 & 14.94$\pm$0.0 & -- & -- & -- & -- & 85.95$\pm$0.0 & 93.61$\pm$0.0 \\
MotionBERT & 73.74$\pm$0.5 & 3.35$\pm$0.0 & 142.61$\pm$1.1 & 6.47$\pm$0.1 & 66.77$\pm$1.0 & 54.96$\pm$1.6 & 66.28$\pm$2.2 & 52.37$\pm$1.1 & -- & -- & -- & -- & 77.77$\pm$0.4 & 92.89$\pm$0.4 \\
Offline Enc-Dec & 76.91$\pm$1.3 & 3.49$\pm$0.1 & 145.14$\pm$1.5 & 6.59$\pm$0.1 & 66.34$\pm$0.9 & 50.61$\pm$2.9 & 64.20$\pm$2.1 & 50.56$\pm$0.4 & -- & -- & -- & -- & 76.78$\pm$0.3 & 92.69$\pm$0.3 \\
\midrule
\multicolumn{15}{l}{\textit{Controlled contextual baselines (identical VLM interface)}} \\
Single-stream & 92.26$\pm$1.8 & 4.19$\pm$0.1 & 146.98$\pm$0.5 & 6.67$\pm$0.0 & 45.32$\pm$8.0 & 13.70$\pm$9.4 & 60.43$\pm$1.4 & 18.08$\pm$5.4 & 65.52$\pm$0.0 & 26.39$\pm$0.0 & 53.86$\pm$0.0 & 14.00$\pm$0.0 & 59.19$\pm$1.1 & 81.57$\pm$0.5 \\
Late fusion & 79.93$\pm$2.4 & 3.63$\pm$0.1 & 140.89$\pm$2.1 & 6.39$\pm$0.1 & 68.42$\pm$1.2 & 53.02$\pm$7.5 & 72.25$\pm$3.9 & 54.23$\pm$8.0 & 92.45$\pm$0.3 & 88.23$\pm$0.9 & 82.92$\pm$0.2 & 67.99$\pm$2.3 & 65.98$\pm$1.0 & 86.04$\pm$0.7 \\
Cross-Attn only & 79.27$\pm$0.9 & 3.60$\pm$0.0 & 141.23$\pm$1.0 & 6.41$\pm$0.1 & 66.61$\pm$2.1 & 61.31$\pm$1.6 & 77.45$\pm$1.9 & 69.27$\pm$1.9 & 92.12$\pm$0.7 & 88.18$\pm$0.7 & 83.36$\pm$0.4 & 68.72$\pm$2.0 & 66.75$\pm$0.4 & 85.85$\pm$0.1 \\
\midrule
\multicolumn{15}{l}{\textit{Driver-WM variants}} \\
\rowcolor{gray!10}
\textbf{Driver-WM (main)} & 71.66$\pm$0.3 & 3.25$\pm$0.0 & 138.34$\pm$1.2 & 6.28$\pm$0.1 & 73.34$\pm$1.3 & 69.09$\pm$1.8 & 80.35$\pm$0.8 & 73.08$\pm$1.0 & 92.94$\pm$0.8 & 89.26$\pm$1.3 & 82.81$\pm$0.4 & 66.54$\pm$1.6 & 71.40$\pm$0.3 & 88.21$\pm$0.3 \\
KL bottleneck & 71.53$\pm$0.9 & 3.25$\pm$0.0 & 137.85$\pm$1.4 & 6.26$\pm$0.1 & 73.89$\pm$1.9 & 67.43$\pm$2.4 & 78.38$\pm$1.4 & 71.08$\pm$1.2 & 92.67$\pm$0.2 & 89.13$\pm$0.5 & 82.27$\pm$1.4 & 66.40$\pm$3.5 & 71.17$\pm$0.4 & 88.31$\pm$0.1 \\
+Skeleton-Pre & 72.32$\pm$1.3 & 3.28$\pm$0.1 & 138.35$\pm$2.0 & 6.28$\pm$0.1 & 72.58$\pm$1.3 & 68.04$\pm$2.9 & 80.51$\pm$1.4 & 73.98$\pm$1.7 & 92.77$\pm$0.9 & 89.01$\pm$1.1 & 81.88$\pm$2.4 & 69.57$\pm$0.7 & 70.93$\pm$1.1 & 87.94$\pm$0.3 \\
RSSM/GRU dyn. & 76.48$\pm$1.7 & 3.47$\pm$0.1 & 137.21$\pm$1.1 & 6.23$\pm$0.1 & 71.53$\pm$0.7 & 67.70$\pm$1.4 & 79.86$\pm$0.9 & 72.29$\pm$1.5 & 92.34$\pm$0.2 & 87.74$\pm$0.5 & 81.50$\pm$0.9 & 63.91$\pm$0.8 & 68.46$\pm$0.9 & 86.97$\pm$0.3 \\
w/o ext context & 71.74$\pm$0.0 & 3.26$\pm$0.0 & 136.94$\pm$0.0 & 6.21$\pm$0.0 & 73.40$\pm$0.0 & 71.33$\pm$0.0 & 74.22$\pm$0.0 & 63.56$\pm$0.0 & 65.52$\pm$0.0 & 26.39$\pm$0.0 & 53.86$\pm$0.0 & 14.00$\pm$0.0 & 71.13$\pm$0.0 & 88.20$\pm$0.0 \\
\bottomrule
\end{tabular}%
}
\end{table*}

\begin{table*}[t]
\centering
\caption{\textbf{Reference semantic comparison on AIDE (test set).}
We report task-wise \textbf{Accuracy} (Acc, \%) for the four AIDE semantic tasks.
\textbf{Note:} classification-only SOTAs are purely discriminative and \emph{cannot} perform multi-step kinematic rollouts; therefore, kinematic metrics (MPJPE/PCK) are inapplicable.
AIDE names the 4th task as \textbf{VCR} (Vehicle Condition Recognition), while several later works denote the same task as \textbf{VBR} (Vehicle Behavior Recognition) for the same label set.
}
\label{tab:supp_sota_acc}
\small
\setlength{\tabcolsep}{4pt}
\renewcommand{\arraystretch}{1.05}
\resizebox{\textwidth}{!}{%
\begin{tabular}{l c c c c c c}
\toprule
\textbf{Model} & \textbf{Paradigm} & \textbf{Rollout} &
\textbf{DBR Acc} & \textbf{DER Acc} & \textbf{TCR Acc} & \textbf{VCR/VBR Acc} \\
\midrule
\multicolumn{7}{l}{\textit{Classification-only SOTAs (no kinematic rollout)}} \\
UV-M3TL~\cite{liu2026uvm3tl} & Classif. & \textbf{$\times$}$^{\dagger}$ & 73.82 & 77.39 & 96.57 & 87.07 \\
TEM$^{3}$-Learning~\cite{09d0b52791304de985947c5608595615} & Classif. & \textbf{$\times$}$^{\dagger}$ & 69.31 & 75.00 & 96.29 & 86.11 \\
MMTL-UniAD~\cite{Liu_2025_CVPR} & Classif. & \textbf{$\times$}$^{\dagger}$ & 73.61 & 76.67 & 93.91 & 85.00 \\
\midrule
\multicolumn{7}{l}{\textit{Driver-WM (ours): kinematic rollout + auxiliary semantics}} \\
\rowcolor{gray!10}
Driver-WM (seed 42) & Rollout+Aux & \textbf{$\checkmark$} & 73.23 & 79.47 & 93.10 & 82.59 \\
Driver-WM (3 seeds, mean$\pm$std) & Rollout+Aux & \textbf{$\checkmark$} & 73.34$\pm$1.31 & 80.35$\pm$0.83 & 92.94$\pm$0.75 & 82.81$\pm$0.38 \\
\bottomrule
\end{tabular}%
}
\vspace{0.6mm}

\begin{minipage}{\textwidth}
\tiny
$^{\dagger}$ Classification-only methods are optimized for clip-level recognition and do not generate future kinematics; MPJPE/PCK and test-time rollout interventions are not applicable.
\end{minipage}
\end{table*}

\clearpage
\section{Direct-GT Intervention Accuracy}
\label{supp:direct_gt_intervention}

The main paper reports intervention-induced deviations from the factual rollout to characterize mechanism sensitivity. Here we complement that analysis by evaluating each intervention directly against the ground-truth skeletons, without retraining. This separates sensitivity from predictive accuracy: a useful intervention probe should not only change the rollout, but should also reveal whether the factual external-to-internal pathway is more accurate than altered pathways.

\begin{table}[t]
\centering
\caption{\textbf{Direct-GT intervention accuracy on the High-Motion subset at $h{=}5$.}
Errors are measured against ground-truth skeletons; lower is better.}
\label{tab:supp_direct_gt_intervention}
\small
\setlength{\tabcolsep}{5pt}
\begin{tabular}{lcc}
\toprule
\textbf{Setting} & \textbf{HM $h{=}5$ MPJPE}$\downarrow$ & \textbf{HM hands $h{=}5$}$\downarrow$\\
\midrule
Factual & 155.82 & 184.63\\
$do(\mathrm{Ext}{=}\emptyset)$ & 164.48 & 199.08\\
$do(g_t{=}0)$ & 226.55 & 265.14\\
$do(g_t{=}1)$ & 170.93 & 205.13\\
$do(\lambda_{\rm CA}{=}2)$ & 180.21 & 206.33\\
\bottomrule
\end{tabular}
\end{table}

Removing external context, disabling the injection pathway, forcing full injection, and over-injecting all increase the direct-GT error relative to the factual rollout. The degradation is especially pronounced on hand joints, supporting the role of the learned, time-varying external-to-internal injection in reactive driver-motion forecasting. For $g_t\in\{0,1\}$, gate clamping is equivalent to the scalar override $\lambda_{\rm CA}\in\{0,1\}$; $\lambda_{\rm CA}{=}2$ is used only as an over-injection stress test.

\section{Oracle External-Latent Diagnostic}
\label{supp:oracle_ext}

We further audit whether external-latent drift is a dominant bottleneck in the current $5{\rightarrow}5$ setting. Under causal-prefix decoding, replacing the future external suffix leaves the current update invariant (max abs diff $=0$), confirming zero-lookahead behavior. Eval-only oracle latents do not improve HM $h{=}5$ MPJPE (155.82 factual vs. 157.85 oracle). A non-deployable Oracle-Ext variant with future external latents teacher-forced during training and evaluation reaches 152.19 HM $h{=}5$ MPJPE. The small upper-bound gain suggests that the remaining gap mainly lies in internal response dynamics and coupling calibration rather than external-latent drift.

\begin{table}[t]
\centering
\caption{\textbf{Oracle external-latent diagnostic on HM $h{=}5$ MPJPE.}}
\label{tab:supp_oracle_ext}
\small
\begin{tabular}{lc}
\toprule
\textbf{Setting} & \textbf{HM $h{=}5$ MPJPE}$\downarrow$\\
\midrule
Driver-WM factual & 155.82\\
Eval-only oracle latents & 157.85\\
Oracle-Ext train/eval teacher forcing & 152.19\\
\bottomrule
\end{tabular}
\end{table}

\clearpage
\section{Safety-Relevant Risk-Ranking Proxy}
\label{supp:risk_ranking_proxy}

We evaluate whether Driver-WM outputs can form a lightweight risk-ranking signal for future high-motion driver reactions. Positives are GT-HM events, defined as the top 10\% test clips ranked by future ground-truth motion. The score uses only Driver-WM outputs: predicted skeleton motion, DBR/DER driver-state probabilities, and seed-variance uncertainty. Each component is normalized using validation-set statistics and then applied to the test set.

\begin{table}[t]
\centering
\caption{\textbf{Safety-relevant risk-ranking proxy on GT-HM events.}
GT-HM is a top-10\% event, so random AUPRC is approximately 0.10.}
\label{tab:supp_risk_ranking}
\small
\setlength{\tabcolsep}{5pt}
\begin{tabular}{lccc}
\toprule
\textbf{Risk score} & \textbf{AUROC}$\uparrow$ & \textbf{AUPRC}$\uparrow$ & \textbf{R@10\%FPR}$\uparrow$\\
\midrule
Motion only & 0.6325 & 0.1364 & 0.1667\\
+ driver semantics & 0.6511 & 0.1567 & 0.1667\\
+ uncertainty & 0.6573 & 0.1603 & 0.2000\\
+ driver semantics + uncertainty & \textbf{0.6726} & \textbf{0.1785} & \textbf{0.2167}\\
\bottomrule
\end{tabular}
\end{table}

The composite score improves AUPRC from 0.1364 to 0.1785 over motion-only. A paired bootstrap test gives a positive gain with 95\% CI $[0.013,0.094]$ and $p{=}0.0009$. This analysis is a risk-ranking proxy rather than a closed-loop planner evaluation.

\clearpage

\section{Additional Post-hoc Visualizations}
\label{app:renderer_vis}

\begin{figure}[ht]
    \centering
    \includegraphics[width=\textwidth]{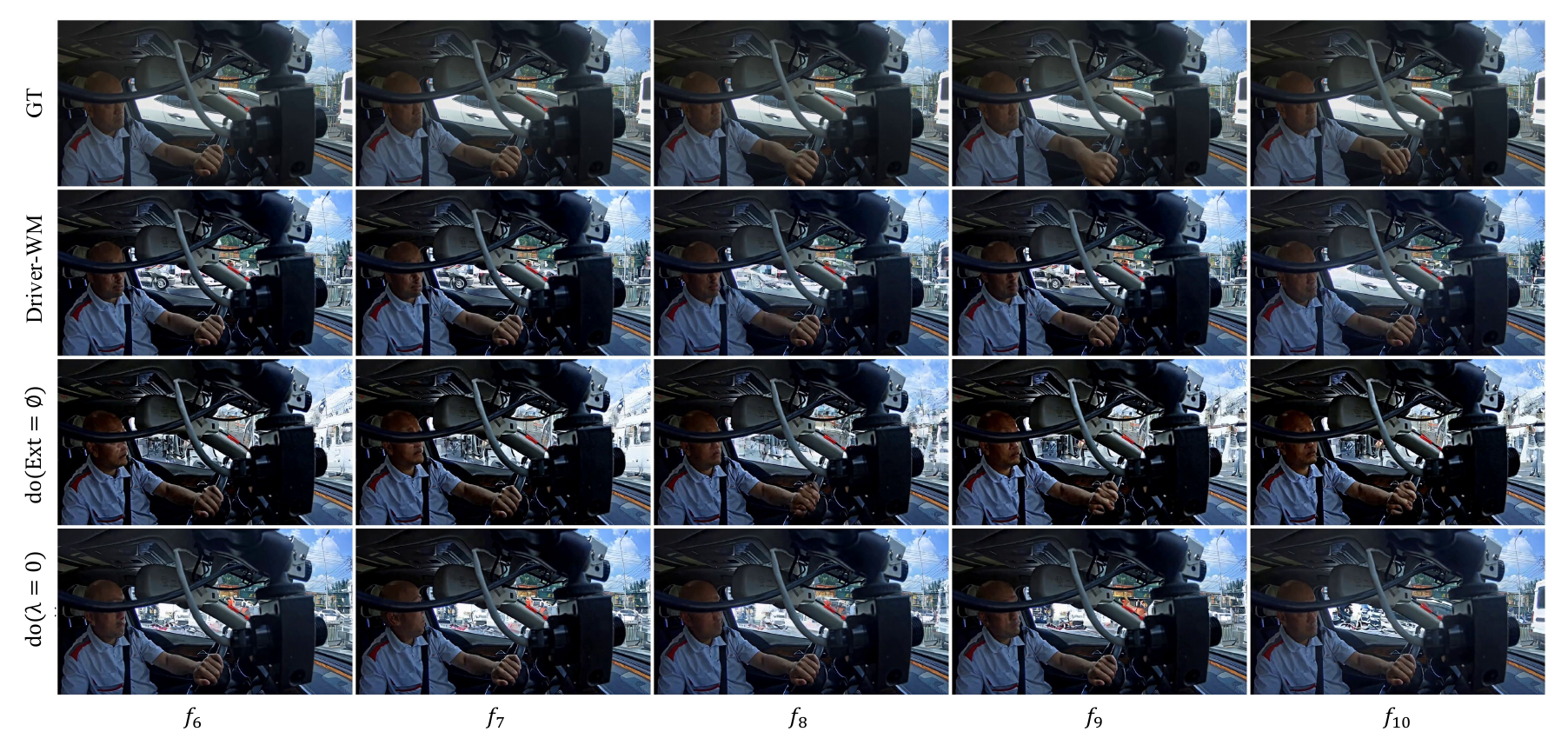}
    \caption{\textbf{Additional post-hoc visualizations with the optional frozen renderer.}
    Five uniformly sampled future frames are shown for one representative test clip under the $5{\rightarrow}5$ rollout protocol. Rows correspond to ground truth (GT), the factual rollout of Driver-WM, and two interventions: $do(\mathrm{Ext}{=}\emptyset)$ and $do(\lambda_{\mathrm{CA}}{=}0)$. The renderer is used only for qualitative interpretation and is not part of training or evaluation. Compared with the factual rollout, the two interventions induce visibly different head and upper-body motion trends, consistent with the skeleton-level intervention analysis reported in the main paper.}
    \label{fig:supp_render}
\end{figure}

For qualitative interpretation only, we additionally visualize selected rollouts using the optional frozen renderer illustrated in the main paper. This renderer is an off-the-shelf post-hoc visualization utility and is not part of the reported training, evaluation, or model selection pipeline. Accordingly, all quantitative results in the main paper and this supplement are computed strictly from skeleton rollout metrics and auxiliary semantic predictions, rather than from rendered pixel-space appearance.

Figure~\ref{fig:supp_render} shows one representative clip under the same $5{\rightarrow}5$ rollout protocol used throughout the paper. Columns correspond to five uniformly sampled future frames across the prediction window, and rows compare the ground truth (GT), the factual rollout of Driver-WM, and two intervened rollouts: $do(\mathrm{Ext}{=}\emptyset)$ and $do(\lambda_{\mathrm{CA}}{=}0)$. These rendered frames are provided solely to improve visual interpretability of the predicted driver motion under different contextual conditions.

The renderer reveals clear differences in the temporal trend of driver responses under different conditions. In this example, the factual rollout remains comparatively stable and closer to the ground-truth future sequence. In contrast, the two interventions induce visibly different head and upper-body responses from the second sampled frame onward: removing external context leads to a more evident upward head motion, while disabling the contextual pathway induces a stronger head-turning tendency. These rendered frames serve as a qualitative counterpart to the skeleton-level intervention analysis reported in the main paper.

\section{Notation and Definitions}
\label{sec:notation_table}

Table~\ref{tab:notation} summarizes key symbols and conventions.
Bold lowercase letters denote vectors and bold uppercase letters denote matrices.
Subscripts indicate time indices. Hat $\hat{\cdot}$ denotes predictions (rollouts).

\begingroup
\small
\renewcommand{\arraystretch}{1.2}
\begin{longtable}{@{} l p{0.75\textwidth} @{}}
\caption{\textbf{Summary of key notations and definitions.}}
\label{tab:notation} \\
\toprule
\textbf{Symbol} & \textbf{Description} \\
\midrule
\endfirsthead

\multicolumn{2}{@{}l}{\tablename~\thetable\ \textit{(continued from previous page)}} \\
\toprule
\textbf{Symbol} & \textbf{Description} \\
\midrule
\endhead

\bottomrule
\endfoot

\bottomrule
\endlastfoot

\multicolumn{2}{c}{\textit{Time and sequences}} \\
\midrule
$t \in \{1,\dots,T\}$ & Time index. \\
$T_{\mathrm{obs}},\,T_{\mathrm{pred}},\,T$ & Observed steps, predicted steps, and total horizon ($T=T_{\mathrm{obs}}+T_{\mathrm{pred}}$). \\
$\mathbf{x}_{1:T}=\{\mathbf{x}_t\}_{t=1}^{T}$ & A generic sequence over time. \\
$\hat{\cdot}$ & Prediction/rollout operator, e.g., $\hat{\mathbf{x}}_t$. \\
\midrule
\multicolumn{2}{c}{\textit{Observations}} \\
\midrule
$\mathbf{o}^{\text{in}}_{t}$, $\mathbf{o}^{\text{out}}_{t}$ & In-cabin and out-cabin observations at time $t$ (out-cabin can be multi-view). \\
\midrule
\multicolumn{2}{c}{\textit{Perception backbone and features}} \\
\midrule
$E_{\text{vlm}}(\cdot)$ & Frozen VLM visual encoder. \\
$\mathbf{f}^{\text{in}}_{t}\in\mathbb{R}^{D}$ & In-cabin feature: $\mathbf{f}^{\text{in}}_{t}=E_{\text{vlm}}(\mathbf{o}^{\text{in}}_{t})$. \\
$\mathbf{f}^{\text{out}}_{t}\in\mathbb{R}^{D}$ & Out-cabin feature: $\mathbf{f}^{\text{out}}_{t}=E_{\text{vlm}}(\mathbf{o}^{\text{out}}_{t})$ (per-view if applicable). \\
$\mathbf{e}_{\text{view}}(v)\in\mathbb{R}^{D}$ & Learnable view embedding for view identifier $v$. \\
$\bar{\mathbf{f}}_{t}$ & View-conditioned feature: $\bar{\mathbf{f}}_{t}=\mathbf{f}_{t}+\mathbf{e}_{\text{view}}(v)$. \\
$A_{\phi}(\cdot)$ & Optional residual adapter; default interface is identity. \\
\midrule
\multicolumn{2}{c}{\textit{Dual-stream latent states}} \\
\midrule
$\mathbf{z}^{\text{int}}_{t}\in\mathbb{R}^{D}$ & Internal (driver/cabin) latent state at time $t$. \\
$\mathbf{Z}^{\text{ext}}_{t}=\{\mathbf{z}^{\text{ext}}_{t,v}\}_{v=1}^{V}$ & Multi-view external latents at time $t$; $v$ indexes views and $V$ is the total number. \\
$\bar{\mathbf{z}}^{\text{ext}}_{t}$ & Mean-pooled external latent: $\bar{\mathbf{z}}^{\text{ext}}_{t}=\frac{1}{V}\sum_{v=1}^{V}\mathbf{z}^{\text{ext}}_{t,v}$. \\
(Default interface) & $\mathbf{z}^{\text{int}}_{t}=\bar{\mathbf{f}}^{\text{in}}_{t}$ and $\mathbf{z}^{\text{ext}}_{t,v}=\bar{\mathbf{f}}^{\text{out}}_{t,v}$. \\
$\hat{\mathbf{z}}^{\text{int}}_{t}$ & Rolled-out (predicted) internal latent at time $t$. \\
$\hat{\mathbf{Z}}^{\text{int}}_{\le t}$, $\bar{\mathbf{Z}}^{\text{ext}}_{\le t}$ & Stacked histories up to time $t$: $\hat{\mathbf{Z}}^{\text{int}}_{\le t}=\{\hat{\mathbf{z}}^{\text{int}}_{\tau}\}_{\tau\le t}$, $\bar{\mathbf{Z}}^{\text{ext}}_{\le t}=\{\bar{\mathbf{z}}^{\text{ext}}_{\tau}\}_{\tau\le t}$. \\
\midrule
\multicolumn{2}{c}{\textit{Time-causal interaction and gated coupling}} \\
\midrule
$\mathrm{Ctx}_{\theta}(\cdot)$ & Time-causal interaction operator (instantiated as causal cross-attention). \\
$\mathbf{m}_{t}\in\mathbb{R}^{D}$ & External-to-internal context summary: $\mathbf{m}_{t}=\mathrm{Ctx}_{\theta}(\hat{\mathbf{Z}}^{\text{int}}_{\le t},\bar{\mathbf{Z}}^{\text{ext}}_{\le t})$. \\
$f_{\theta}(\cdot)$ & Internal transition predictor producing $(\boldsymbol{\mu}_{t+1},\log\boldsymbol{\sigma}_{t+1})$ from current internal latent. \\
$\tilde{\mathbf{z}}^{\text{int}}_{t+1}$ & Candidate next internal latent via reparameterization: $\tilde{\mathbf{z}}^{\text{int}}_{t+1}=\boldsymbol{\mu}_{t+1}+\boldsymbol{\epsilon}\odot\boldsymbol{\sigma}_{t+1}$ (deterministic case uses $\boldsymbol{\mu}_{t+1}$). \\
$\mathbf{g}_{t}\in(0,1)^{D}$ & Vector-valued gate: $\mathbf{g}_{t}=\sigma(\mathrm{MLP}_g(\bar{\mathbf{z}}^{\text{ext}}_{t}))$. \\
$\hat{\mathbf{z}}^{\text{int}}_{t+1}$ & Gated causal update: $\hat{\mathbf{z}}^{\text{int}}_{t+1}=(\mathbf{1}-\mathbf{g}_t)\odot\tilde{\mathbf{z}}^{\text{int}}_{t+1}+\mathbf{g}_t\odot\mathbf{m}_t$. \\
\midrule
\multicolumn{2}{c}{\textit{Outputs and heads}} \\
\midrule
$K$ & Number of skeleton keypoints (e.g., $K{=}136$ for HALPE-136). \\
$D_{\text{skel}}(\cdot)$ & Skeleton decoding head. \\
$\hat{\mathbf{s}}_{t}$ & Predicted 2D skeleton: $\hat{\mathbf{s}}_{t}=D_{\text{skel}}(\hat{\mathbf{z}}^{\text{int}}_{t}) \in \mathbb{R}^{K\times 2}$. \\
$D_{\text{dbr}},D_{\text{der}}$ & Auxiliary semantic heads attached to internal latents. \\
$D_{\text{tcr}},D_{\text{vcr}}$ & Auxiliary semantic heads attached to external latents. \\
\midrule
\multicolumn{2}{c}{\textit{Optional probabilistic bottleneck}} \\
\midrule
$q_{\theta}(\tilde{\mathbf{z}}^{\text{int}}_{t+1}\mid \hat{\mathbf{z}}^{\text{int}}_{t})$ & Diagonal Gaussian transition $\mathcal{N}(\boldsymbol{\mu}_{t+1},\mathrm{diag}(\boldsymbol{\sigma}_{t+1}^2))$. \\
$\mathcal{L}_{\text{KL}}$ & Supervised KL regularizer (ablation only). Main model is deterministic. \\
\midrule
\multicolumn{2}{c}{\textit{Training objectives}} \\
\midrule
$\mathcal{L}_{\text{skel}}$ & Skeleton regression loss on $\{\hat{\mathbf{s}}_t\}$. \\
$\mathcal{L}_{\text{phys}}$ & Physical priors (bone-length, temporal smoothness, ROI feasibility). \\
$\mathcal{L}_{\text{lat}}$ & Latent rollout consistency loss. \\
$\mathcal{L}_{\text{aux}}$ & Auxiliary semantic supervision losses. \\
$\mathcal{L}$ & Total objective: $\mathcal{L}=\lambda_{\text{lat}}\mathcal{L}_{\text{lat}}+\lambda_{\text{skel}}\mathcal{L}_{\text{skel}}+\lambda_{\text{aux}}\mathcal{L}_{\text{aux}}+\lambda_{\text{phys}}\mathcal{L}_{\text{phys}}$. \\
\end{longtable}
\endgroup
% \input{supp_eccv2026}
% ------------

\end{document}